\documentclass[letterpaper, 10 pt, conference]{ieeeconf}  
\IEEEoverridecommandlockouts                              

\usepackage{soul}
\usepackage{cite}
\usepackage{amsmath,amssymb,amsfonts}
\usepackage{siunitx}
\usepackage{graphicx}
\usepackage{xcolor}
\usepackage{hyperref}
\usepackage{svg}
\usepackage{cases}
\usepackage{subcaption}
\usepackage{derivative}
\usepackage[capitalise]{cleveref}
\usepackage[linesnumbered,ruled,vlined]{algorithm2e}
\usepackage{algpseudocode}
\usepackage{geometry}
\usepackage{fancyhdr}
\usepackage{booktabs}

\usepackage{hyperref}

\hypersetup{
    colorlinks=true,
    urlcolor=blue
}
\SetKwInput{KwInput}{Input}                
\SetKwInput{KwOutput}{Output}              

\newtheorem{theorem}{Theorem}

\AtBeginDocument{
    \newgeometry{
        top=57pt,
        left=48pt,
        right=48pt,
        bottom=43pt,
        includehead,
        includefoot
    }
}

\newcommand{\disable}[1]{}

\title{\LARGE \bf
OGM-CBF: Occupancy Grid Map-based Control Barrier Function for Safe Mobile Robot Control with Memory of out of View Obstacles}

\author{%
  Golnaz Raja, Miloš Prágr, Topi R. J. Kärki, Teemu Mökkönen, Reza Ghabcheloo%
\thanks{%
The authors are with the Faculty of Engineering and Natural Sciences, Tampere University, Kalevantie 4, 33100 Tampere, Finland;
{\texttt \{golnaz.raja, milos.pragr, topi.karki, teemu.mokkonen, reza.ghabcheloo\}@tuni.fi};
corresponding author: Golnaz Raja.
This work has been supported by Finland's Ministry of Education and Culture’s Doctoral Education Pilot under Decision No. VN/3137/2024-OKM-6, 
by European Union's Horizon Europe under Grants 101095947 and 101136408.
}%
}

\begin{document}

\onecolumn
\pagenumbering{roman}
\thispagestyle{empty}

{\LARGE\bfseries{}IEEE Copyright Notice\par}

\vspace{0.8cm}

{\large
\noindent
© 2026 IEEE.
Personal use of this material is permitted.
Permission from IEEE must be obtained for all other uses, in any current or future media, 
including reprinting/republishing this material for advertising or promotional purposes, 
creating new collective works, for resale or redistribution to servers or lists,
 or reuse of any copyrighted component of this work in other works.

}

\vspace{0.8cm}

{\large
\noindent
This work has been accepted for publication in IEEE/RSJ International Conference on Intelligent Robots and Systems (IROS) - September 27 - October 1, 2026.
The final published version will be available via IEEE Xplore.
\par
}

\vspace{0.8cm}

{\large
\noindent DOI: \textit{TBD}
\par
}

\clearpage
\twocolumn
\setcounter{page}{1}
\pagenumbering{arabic}


\maketitle

\begin{abstract}
Safe control in unknown environments is a key challenge in mobile robotics. Control Barrier Functions (CBFs) provide a principled framework for guaranteeing safety constraint satisfaction. State-of-the-art CBF approaches assume either known environments with predefined obstacles, 
or rely only on obstacles currently within the robot’s Field of View (FoV). However, practical robots in a priori unknown environments can observe their surroundings only partially, and therefore can violate safety due to limited FoV, sensor range, or occlusion. This paper incorporates the memory of previously observed obstacles of arbitrary shape that have left the robot's FoV into CBF-based safe control.
In particular, we couple the Signed Distance Function (SDF)-based CBF formulation to an occupancy grid map built online during the system's operation. Furthermore, the lack of steering authority induced by the SDF gradient degeneracy when facing obstacles head-on is addressed by constructing a Gaussian pyramid of the SDF, yielding a multi-level CBF.
The efficacy of the proposed approach is evaluated against memory unaware baselines in the CARLA simulator. Moreover, we demonstrate the generalizability of the proposed approach in real deployments on a small warehouse robot and a large, articulated frame steering autonomous wheel loader. Project website: \href{https://ogmcbf.github.io/}{ogmcbf.github.io}
\end{abstract}
\color{black}
\color{black}


\newcommand{\state}{\ensuremath{\boldsymbol{x}}}
\newcommand{\control}{\ensuremath{\boldsymbol{u}}}
\newcommand{\position}{\ensuremath{\boldsymbol{p}}}
\newcommand{\heading}{\ensuremath{\boldsymbol{\hat{\position}}}}
\newcommand{\observation}{\ensuremath{\boldsymbol{z}}}
\newcommand{\envState}{\ensuremath{\boldsymbol{q}}}

\newcommand{\cmdlin}{\ensuremath{\boldsymbol{v}}}
\newcommand{\cmdang}{\ensuremath{\boldsymbol{\omega}}}

\newcommand{\stateSpace}{\ensuremath{\mathbb{X}}}
\newcommand{\stateSpaceSafe}{\ensuremath{\stateSpace_\text{s}}}
\newcommand{\stateSpaceUnsafe}{\ensuremath{\stateSpace_\text{u}}}
\newcommand{\controlSpace}{\ensuremath{\mathbb{U}}}
\newcommand{\controlSpaceSafe}{\ensuremath{\controlSpace_\text{s}}}

\newcommand{\superlevel}{\ensuremath{\mathcal{S}}}
\newcommand{\cbf}{\ensuremath{h}}
\newcommand{\cbfAlpha}{\ensuremath{\alpha}}

\newcommand{\gridmap}{\ensuremath{\mathcal{M}}}
\newcommand{\cell}{\ensuremath{\nu}}
\newcommand{\cellsize}{\ensuremath{d_{\cell}}}
\newcommand{\occupancy}{\ensuremath{\text{prob}}}
\newcommand{\occupancyBinary}{\ensuremath{\occupancy_\text{bin}}}
\newcommand{\occupancyThr}{\ensuremath{\occupancy_\text{occ}}}

\newcommand{\sdf}{\ensuremath{\phi}}
\newcommand{\sdfSmooth}{\ensuremath{\phi'}}
\newcommand{\hyperS}{\ensuremath{l_\text{s}}}
\newcommand{\hyperA}{\ensuremath{l_\text{a}}}

\section{Introduction}

Autonomous vehicles, mobile industrial robots, and multi-agent systems need to respect safety requirements when navigating complex unknown environments. Control Barrier Functions (CBFs)~\cite{ames2019control} provide strong mathematical guarantees of satisfying the safety requirements, with promising applications on legged robots \cite{nguyen20163d}, autonomous vehicles \cite{ames2014control}, multi-robot systems \cite{wang2017safety}, and fast aerial robots~\cite{singletary2022onboard}. CBFs can be combined with Control Lyapunov Functions (CLFs) in Quadratic Programs (QPs) to ensure safety and stability in safety-critical systems \cite{ames2016control}, or used as QP-based safety filters that prevent safety violations \cite{ames2019control} by minimal modifying nominal controls from operators, classical planners (e.g., MPC), or learned policies.

State-of-the-art CBFs define safe sets by reasoning about explicitly identified obstacles or build Signed Distance Functions (SDFs) of safe sets.
BarrierNet~\cite{xiao2023barriernet} optimizes parameters of a CBF constructed from sensor data with respect to (w.r.t.) user-defined safe sets, while Liu et al.~\cite{liu2023clf} extract elliptical obstacles from elevation grid maps. However, approximating obstacles with minimum bounding circles requires a separate constraint for each obstacle, and is overly conservative for some environments.
SDFs provide a representation of the closest distance to a set's surface, offering essential information for online obstacle avoidance and motion planning~\cite{han2019fiesta}.

Li et al.~\cite{li_collision-free_2025} formulate an SDF-based CBF to maneuver around obstacles in a goal reaching scenario using local sensor measurements.
Srinivasan et al.~\cite{srinivasan2020synthesis} learn to classify safe sets in LiDAR data using a Support Vector Machine and construct a single constraint CBF from the resulting SDF at the cost of exhaustive data collection.
In~\cite{long2021learning}, the computational complexity of~\cite{srinivasan2020synthesis} is addressed by approximating the SDF from LiDAR using an incremental training approach with replay memory, albeit employing one constraint per obstacle.

\begin{figure}[tb]
\centering
\includegraphics[width=0.7\columnwidth]{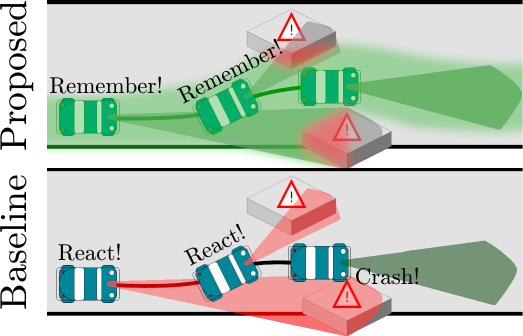}
\caption{%
  (top) The proposed approach exploits a grid map to retain the memory of an obstacle observed a priori, and thus can safely avoid it even after the obstacle leaves the Field of View (FoV).
  (bottom) A baseline approach without memory is unaware of the a priori observed obstacle after it leaves the FoV, resulting in unsafe behavior.
  }
\label{fig:intro}
\vspace{-0.75em}
\end{figure}
In a priori unknown environments, CBFs must rely on awareness of the surrounding unsafe regions and their changes to reach a quick and smooth generation of a safety certificate.
However, practical mobile robots are hindered by the occlusion, range, and Field of View (FoV) of the available sensors such as cameras or solid state LiDARs.
Therefore, at any time, a robot can only partially observe the environment, limiting its ability to reason about safety.
Occupancy grid maps (OGMs) aggregate such partial observations to construct grid-based occupancy models where each cell carries the probability that its respective part of the environment is an obstacle~\cite{moravec1985high}.
OGMs are sensor-agnostic~\cite{li2018high,nguyen2011stereo,wei2023surroundocc}, and can be updated online during the deployment, utilizing a probabilistic approach to account for sensor uncertainty~\cite{collins2007occupancy}.

We propose OGM-CBF, a novel CBF formulation that exploits the OGM's persistency to retain memory of a priori observed obstacles as demonstrated in~\cref{fig:intro}.
The proposed CBF employs smoothly interpolated SDF to describe arbitrarily shaped obstacles into one CBF constraint in total in a QP.
In particular, the CBF has relative degree one, yielding a safety filter over the linear and angular velocities.

We highlight the contributions as follows.
\begin{itemize} 
 
  \item A CBF that employs OGM and smoothly interpolated SDF to efficiently describe obstacles of arbitrary amount and shape using a single CBF constraint in a QP problem.

    \item Integrating OGM's capability to retain memory of a priori observed obstacles that can no longer be observed by the robot into the CBF.

    \item A sensor-agnostic safety guarantee method that has been integrated into a range of exteroceptive modalities such as depth cameras, laser scanners, and 3D LiDARs.
    \item A demonstration on how to employ image pyramid over the SDF to yield a multi-level OGM-CBF formulation and inject steering authority when facing degenerate SDF gradients of head-on approached obstacles.     
    \item Quantitative and qualitative evaluation of the proposed approach against memory-unaware baselines. 
    \item Deployment of the proposed approach on both a small, differential drive warehouse robot and a large, Articulated Frame Steering (AFS) autonomous wheel-loader. 
\end{itemize}

The rest of this paper is organized as follows.
\cref{sec:background} reviews the principles of the CBF, OGM, and SDF.
\cref{sec:method} describes the proposed control methodology.
\cref{sec:results} provides the quantitative and qualitative evaluations, and reports on the deployments on the warehouse robot and wheel loader. Finally, in~\cref{sec:conc}, we conclude the paper and draw the future work.

\section{Preliminaries and Background}
\label{sec:background}

Let the robot's motion be described by the non-linear control-affine system
\begin{equation}\label{eq:background:control_system}
\begin{aligned}
\dot{\state} = f(\state) + g(\state)\control,\\
\observation = O_1(\state,\boldsymbol{\envState}), \\
(\position,\heading) = O_2(\state),
\end{aligned}
\end{equation}
where $\state \in \stateSpace \subseteq \mathbb{R}^{n}$ is the system state, $\control =[\cmdlin,\cmdang]^T \in \controlSpace \subseteq \mathbb{R}^{m}$ is the control input, consisting of the linear and angular velocities in the body frame, and $f : \mathbb{R}^n \to \mathbb{R}^n \quad\text{and}\quad g : \mathbb{R}^n \to \mathbb{R}^{n\times m}$ are locally Lipschitz continuous functions.
The robot is equipped with the observation systems $O_{1}$ and $O_{2}$.
The exteroceptive (e.g., camera, LiDAR, RADAR) $O_{1} : \mathbb{R}^n \times \mathbb{R}^q \rightarrow \mathbb{R}^k$ is a function of the state and the environment variables $\envState \in \mathbb{R}^{q}$ that yields the observation $\observation \in \mathbb{R}^{k}$.
Finally, $O_{2} : \mathbb{R}^n \rightarrow \mathbb{R}^i \times \mathbb{R}^i$ is a function of states and provides the position $\position \in \mathbb{R}^{i}$ and the robot heading unit vector $\heading \in \mathbb{R}^{i}$.

\subsection{Control Barrier Function}

Let the system \eqref{eq:background:control_system} be categorized as safe if $\state(t) \in \stateSpaceSafe \subseteq \stateSpace$, $\control(t)\in \controlSpaceSafe \subseteq \controlSpace$, $\forall t\geq 0$,
where $\stateSpaceSafe$ is the set of safe states, and $\controlSpaceSafe$ is the set of admissible control inputs.
$\superlevel \subseteq \stateSpaceSafe$ is the zero-superlevel set of a continuously differentiable function $\cbf(\state) : \stateSpace \rightarrow \mathbb{R}$, yielding
\begin{equation}\label{eq:background:safe_set}
\superlevel = \{\state \in \mathbb{X} \; | \; \cbf(\state) \geq 0 \}.
\end{equation}

Next, we reason about the safety of~\eqref{eq:background:control_system} in terms of the forward invariant set.
The set $\superlevel$ is forward invariant for the control system \eqref{eq:background:control_system} if the state trajectories starting in $\superlevel$ will remain in $\superlevel$, i.e., if $\state(0) \in \superlevel$, there exists $\control(t) \in \controlSpaceSafe$ such that $\state(t) \in \superlevel$, $\forall t\geq 0$ \cite{blanchini1999set}.
Ames et al.~\cite{ames2019control} define that the function $\cbf(\state)$ is a Control Barrier Function (CBF) if there exists an extended class $\kappa$ function $\cbfAlpha(.)$ such that for all $\state \in \stateSpace$
\begin{equation}
\label{eq:background:dot_cbf}
    \sup_{\control \in \mathbb{U}_s}[\frac{\partial \cbf(\state)}{\partial \state}  (f(\state)+g(\state)\control)] 
    \ge -\cbfAlpha(\cbf(\state)),
\end{equation}
where an extended class $\kappa$ function is a function $\alpha: \mathbb{R}\rightarrow \mathbb{R}$ that is strictly increasing and $\alpha(0)=0$.
Besides, for brevity, let $\Dot{\cbf}(\state,\control)=\frac{\partial \cbf(\state)}{\partial \state}  (f(\state)+g(\state)\control)$ be used further on.

\subsection{Occupancy Grid Map}

Let the environment be represented as the Occupancy Grid Map (OGM) $\gridmap$~\cite{moravec1985high} with cell size $\cellsize$ 
that follows the update rule
\begin{equation}
\label{eq:background:ogm}
  \text{update}(\gridmap, \position,\heading, \observation) \rightarrow \gridmap'.
\end{equation}
Let $\occupancy(\cell)$ be the probability that the position $\position$ corresponding to the cell $\cell(\position)$ is an obstacle. 
The safe and unsafe sets are defined by binarizing $\gridmap$ so that
\begin{equation}\label{eq:background:binary_ogm}
  \occupancyBinary (\cell) =
\begin{cases}
  0 & \text{if}\ \occupancy(\cell) < \occupancyThr,\\
  1 & \text{if}\ \occupancy(\cell) \geq \occupancyThr,
\end{cases}
\end{equation}
where the risk threshold $\occupancyThr$ is the probability at which a cell is considered occupied.
Then, the unsafe set \(\stateSpaceUnsafe\) is the union of all states corresponding to the occupied cells 
\begin{equation}\label{eq:background:unsafe}
  \stateSpaceUnsafe = \{ \state \in \stateSpace \mid (\position,\heading) = O_2(\state) : \occupancyBinary(\cell(\position)) = 1 \},  
\end{equation}
and the safe set $\stateSpaceSafe$ is given by $\stateSpace \setminus \stateSpaceUnsafe$.

\subsection{Signed Distance Function}\label{sec:background:sdf}

Let the Signed Distance Function (SDF) at point $\position$ yield the signed distance from $\position$ to the closest point of a set $\Omega$ in a metric space, where the sign indicates whether $\position$ is inside or outside of $\Omega$~\cite{chan2005level}.
If $\Omega$ is a subset of a metric space with metric $d$, the SDF is
\begin{equation}\label{eq:background:sdf}
\sdf(\position, \Omega)=
\begin{cases}
  d(\position, \partial \Omega)  & \text{if}\ \position \in \Omega,\\
  -d(\position, \partial \Omega) & \text{if}\ \position \notin \Omega,
\end{cases}
\end{equation}
where $\partial \Omega$ is the boundary of $\Omega$ and $d(\position, \partial \Omega)$ is the minimum distance from $\position$ to a point on the boundary.
Additionally, $d(., \partial \Omega)$ is the desired distance metric to the set $\partial \Omega$.
If $\Omega$ is a subset of the Euclidean space $\mathbb{R}^n$ with a piecewise smooth boundary and $d(., \partial \Omega)$ is the Euclidean distance metric, the SDF is differentiable almost everywhere.
Moreover, the gradient of the function satisfies the eikonal equation $|\nabla \sdf| = 1$ at every point where it is differentiable;
for additional details, the interested reader is referred to \cite{dapogny2012computation,luo2019variational}.


\section{Control Methodology}
\label{sec:method}

\begin{figure}[!htb]
\phantom{}
\includegraphics[height=1.45cm]{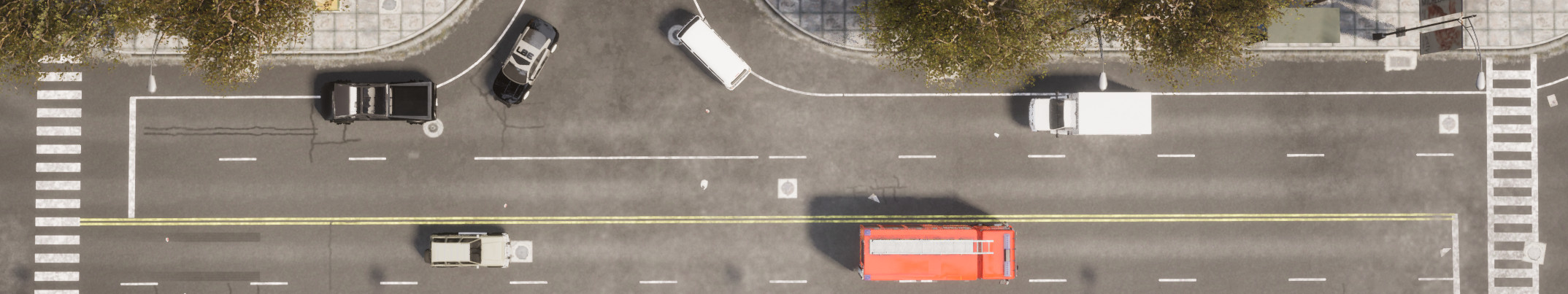}
\hfill
\phantom{}\\[-1.5em]
\subfloat[]{\phantom{\hspace \columnwidth}}\\[0.5em]
\phantom{}
\includegraphics[height=1.7cm]{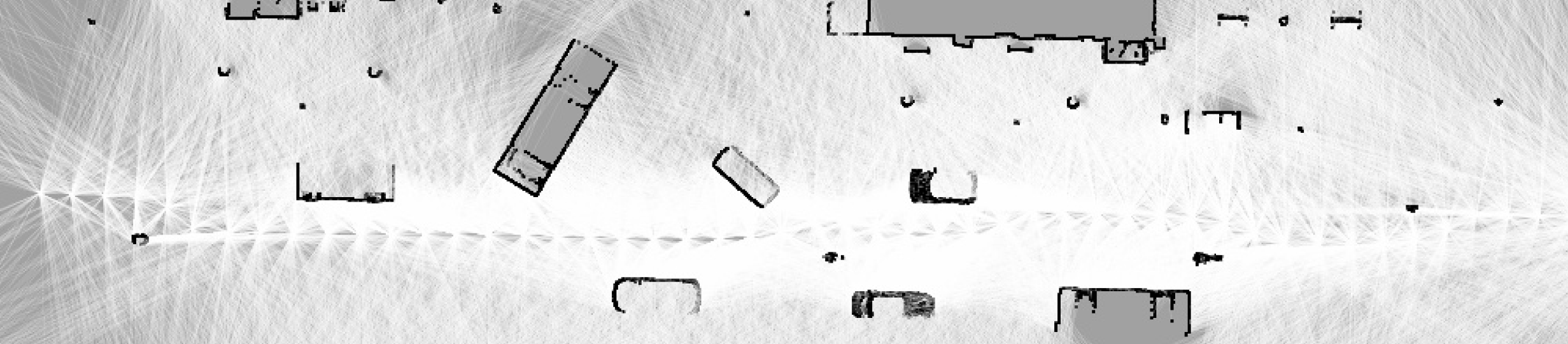}
\hfill
\phantom{}\\[-1.5em]
\subfloat[]{\phantom{\hspace \columnwidth}}\\[0.5em]
\phantom{}
\includegraphics[height=1.7cm]{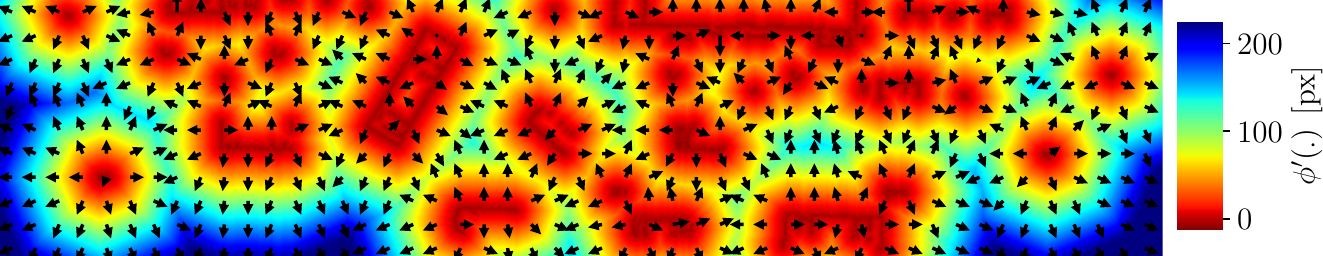}
\hfill
\phantom{}\\[-2.25em]
\subfloat[\label{fig:method:carla_sdf_map:sdf}]{\phantom{\hspace \columnwidth}}\\[0.0em]
\caption{%
  Construction of the smooth SDF $\sdfSmooth$.
  (a) a top-down view of the navigated environment,
  (b) the corresponding OGM, and
  (c) the smooth SDF $\sdfSmooth$ of the safe set and its gradient field.
  }
\label{fig:method:carla_sdf_map}
  \vspace{-1em}
\end{figure}
The proposed method aims to ensure safety of robots in partially observable environments by formulating a CBF that employs OGM to retain memory of a priori observed obstacles that have left the robot's field of view. The CBF comprises an SDF-based distance term that captures the effect of the robot's distance to obstacles of arbitrary shape, and an alignment term that captures the alignment of the robot's heading vector and the avoidance direction.
It is built as follows.

First, the OGM is refined into an SDF of the unoccupied space, see~\cref{fig:method:carla_sdf_map}. Next, the SDF is processed by the image pyramid to retain steering authority when facing the degenerate gradient prevalent in head-ons with locally flat obstacles.
The image pyramid transform yields CBF constraints at multiple scales, providing additional avoidance directions. Besides, as the OGM and thus also the SDF are discrete (in grid space), interpolation is used over each scale SDF to ensure CBF continuity. Finally, following the common safe-control practice, the CBF condition is enforced via a QP-based safety filter that minimally perturbs a nominal performance controller.

In the rest of this section, we describe the individual building blocks in detail.

\subsection{Occupancy Grid Map-based Control Barrier Function}\label{sec:method:ogmcbf}

Let the OGM $\gridmap$ follow the online update rule~\eqref{eq:background:ogm}.
Then, the safety is achieved using the CBF
\begin{equation}\label{eq:method:cbf}
    \cbf(\state) = \sdfSmooth (\position; \stateSpaceSafe) + \hyperS + \hyperA \heading \cdot \nabla_{\position} \sdfSmooth (\position; \stateSpaceSafe),
\end{equation}
where
\begin{itemize}

\item %
  $\sdfSmooth (\position; \stateSpaceSafe) = T(\sdf(\position, \stateSpaceSafe))$, and $T: \mathbb{R}^j \rightarrow \mathbb{R}$ is an interpolation function, where $j$ is the number of points from the OGM used for interpolation. Therefore, $\sdfSmooth$ is a smoothed SDF of safe set $\stateSpaceSafe$, evaluated at the robot's body frame position $\position$;
 \item %
  $\heading \cdot \nabla_{\position} \sdfSmooth (\position; \stateSpaceSafe)$ is the dot product of the robot heading unit vector $\heading$ and smooth SDF gradient $\nabla_{\position} \sdfSmooth (\position;\stateSpaceSafe)$;
\item %
  $\hyperS$ and $\hyperA$ are scalar values such that $0 < \hyperA \leq -\hyperS$.
\end{itemize} 
Note that $\position$, $\heading$, and thus also $\sdfSmooth$ are defined in the same coordinate frame, e.g., the world frame, and $\position$, $\heading$ are derived from system state $\state$ as in \eqref{eq:background:control_system}.

Intuitively, the proposed CBF comprises two distinct components.
The larger these components are, the safer the robot is.
The first component, $\sdfSmooth (\position; \stateSpaceSafe)$, captures the effect of the robot's distance to obstacles, maintaining safety.

The second component, $\hyperS + \hyperA \heading \cdot \nabla_{\position} \sdfSmooth (\position; \stateSpaceSafe)$, captures the alignment of the robot's heading vector $\heading$ to the SDF gradient $\nabla_{\position} \sdfSmooth (\position;\stateSpaceSafe)$.
In particular, the alignment component involves calculating the cosine similarity of the two vectors.
Since $\nabla_{\position} \sdfSmooth$ is the direction in which $\sdfSmooth$ increases the most, moving in that direction moves the robot away from obstacles as visualized by the gradient field in \cref{fig:method:carla_sdf_map:sdf}.
Without the second component, the derivative of $\cbf(\state)$ would be a function of linear velocity only, i.e., to maintain safety, the system could only stop.
Therefore, similar to Toulkani et al.~\cite{toulkani2022reactive}, the second component aims to enable steering without the additional complexity of a higher order CBF such as the one in~\cite{xiao2019control}.

The performance of the OGM-CBF is influenced by the design of the function $\cbfAlpha(.)$, and the values of $\hyperS$ and $\hyperA$, all of which are hyperparameters to be adjusted as per control system~\eqref{eq:background:control_system} to trade-off between the safety margin and responsiveness. The offset $l_s$ shifts the barrier, and $l_a$ increases the sensitivity to robot heading by penalizing motion toward obstacles.

\subsection{Necessary Properties of the Control Barrier Function}\label{app:cbf_proof}

Next, we establish that the constructed function has the necessary properties to be a CBF.

\begin{theorem}
\label{thm:method:1}
Set $\superlevel = \left\{\state\in \stateSpace \; | \; h(\state)\geq 0\right\}$ with \( h(\state) \) defined in \eqref{eq:method:cbf} is a forward-invariant set and \( h(\state) \) is a control barrier function for system \eqref{eq:background:control_system}, with control input $\control= [\boldsymbol{v},\boldsymbol{\omega}]^T$.

\end{theorem}

\begin{proof}
First, we show that $\superlevel\subseteq \stateSpaceSafe$.
As $0<l_{a} \leq -l_{s}$ ensures that $\forall \state$ we have $c \equiv l_{s} + l_{a} \heading \cdot \nabla_{\position} \sdfSmooth (\position; \stateSpaceSafe) \leq 0$,
for any  $\state \in \superlevel=\{\state:h(\state)\ge0\}$, we have $h(\state)\ge0$, therefore $\sdfSmooth (\position; \stateSpaceSafe)= h(\state)-c\ge0$, and consequently $\sdf(\position, \stateSpaceSafe)\ge0$. Thus $\state\in \stateSpaceSafe=\{\state:\sdf(\position, \stateSpaceSafe)\ge0\}$, proving the first part.\\
We will now show that there are control signals $\control$ such that \eqref{eq:background:dot_cbf} is satisfied.
  We will first derive $\dot{h}(\boldsymbol{x},\boldsymbol{u})$, that is
\begin{align}
\label{eq:method:h_dot}
\dot{h} &= \frac{d}{dt} \sdfSmooth (\position; \stateSpaceSafe) 
+ \frac{d}{dt} \left( l_{a} \heading \cdot \nabla_{\position} \sdfSmooth (\position; \stateSpaceSafe) \right),
\end{align}
and assuming that near field obstacles are already integrated in $\mathbb{X}_{s}$ (and the map), far field obstacles have negligible effect on safety, and the underlying environment is stationary once observed, we have
\begin{align}
\label{eq:method:h_dot1}
\dot{h}= \nabla_{\position} \sdfSmooth \cdot \dot{\boldsymbol{p}} + l_{a} \frac{d\heading}{dt}  \nabla_{\position} \sdfSmooth + l_{a}\heading \cdot \frac{d}{dt} \nabla_{\position} \sdfSmooth.
\end{align}

Note that $\heading=R\hat{\boldsymbol{e}}$, where $R$ is the rotation matrix mapping body-frame coordinates to world-frame coordinates, and $\hat{\boldsymbol{e}}=[1,0,0]^\top$ is the heading vector in the body frame. As $\dot{\boldsymbol{p}} = R \boldsymbol{v}$ and  $\frac{d\heading}{d t}=R[\boldsymbol{\omega}]\hat{\boldsymbol{e}}= R(\boldsymbol\omega\times \hat{\boldsymbol e})$, 
substituting into \eqref{eq:method:h_dot1} and using the triple-product identity $(\boldsymbol\omega\times \hat{\boldsymbol e}) \cdot q=\boldsymbol\omega \cdot(\hat{\boldsymbol e}\times q)$ yields

\begin{align}
\label{eq:omega_term_explain_2}
\dot h
&= \underbrace{\Big(R^\top(\nabla_{\position} \sdfSmooth
      + l_a\,  \nabla^2_{\position} \sdfSmooth\ \heading) \cdot \boldsymbol v\Big)}_{a(\boldsymbol{x})\boldsymbol{v}}
\nonumber\\
&\quad + \underbrace{l_a\,\boldsymbol\omega \cdot \Big(\hat{\boldsymbol e}\times (R^\top \nabla_{\position} \sdfSmooth)\Big)}_{b(\boldsymbol{x})\boldsymbol{\omega}}.
\end{align}
  
  Now we have 
  $\dot h=a(\boldsymbol{x})\boldsymbol{v} + b(\boldsymbol{x})\boldsymbol{\omega}$.
Since $(\boldsymbol{v},\boldsymbol{\omega})$ can be made zero, condition \eqref{eq:background:dot_cbf} can be satisfied and thus $h(\boldsymbol{x})$ is a control barrier function for \eqref{eq:background:control_system}, and $\superlevel$ is an invariant set.
\end{proof}

\subsection{SDF Pyramid Transform}
\label{sc.sdf_pyr}
Recall that in (\ref{eq:omega_term_explain_2}), $\boldsymbol{\omega}$ can enter $\dot h$ only through the term $b(\boldsymbol{x})$.
It follows that whenever $\nabla_{\position} \sdfSmooth(\position;\mathbb X_s)$ is collinear with $\heading$, then $b(\state)=0$, and therefore $\dot h$ becomes (locally) insensitive to $\boldsymbol{\omega}$.
The collinearity of $\nabla_{\position} \sdfSmooth(\position;\mathbb X_s)$ with $\heading$ is prevalent when approaching locally orthogonal obstacle boundaries head on. In this \emph{gradient degeneracy} regime, the CBF cannot exploit $\boldsymbol{\omega}$ to steer away and must reduce $\boldsymbol{v}$, losing the steering authority.

We mitigate this by constructing a multi-scale Gaussian pyramid of the SDF with CBF constraints at each scale.
From the base SDF grid $\sdf_1$, we build levels $k=2,\dots,K$ via smoothing and downsampling by the factor of two

\begin{equation}
\sdf_{k} \;=\; \mathcal{D}\!\left(\mathcal{G}_\sigma * \sdf_{k-1}\right),
\end{equation}
where $\mathcal{G}_\sigma$ is the Gaussian blur kernel, convolved at each level, followed by the pyramid down sampling $\mathcal{D}$.

Finally, we define a level-wise barrier $h_k(\state)$ by replacing $\sdf$ and its gradient with $\sdf_k$ and $\nabla\sdf_k$, respectively, for $k=1,\dots,K$, and impose one CBF, $h_k(\state)$, per level. The coarser but smoothed levels reduce local gradient degeneracy and provide additional, non-parallel avoidance directions, improving steering around locally orthogonal obstacle geometries. 

\subsection{Smooth Signed Distance Function Approximation}\label{sec:method:smooth}

Since the occupancy grid map $\gridmap$ is calculated at discrete positions (centers of the cells), $\phi(.; \stateSpaceSafe)$ is initially also calculated at discrete grid points, and thus is not continuous.
However, it is then smoothly interpolated to approximate the SDF at an arbitrary robot position $\position$, using $j$ grid points neighboring $\position$.
Let the function $T$ represent this interpolation (smoothing) of the SDF;
then, applying $T$ ensures the continuity of $\cbf(\state)$.

\subsection{OGM-CBF-QP Control Synthesis}
The safe control using the proposed OGM-CBF is implemented as a QP-based safety filter.
At each control step, given a nominal/reference control input $\control_{\mathrm{ref}}(\state)$, a safe control input is computed by solving
\begin{equation}\label{eq:method:optimization_cbfqp}
\begin{aligned}
    \control^{*}(\state)
    &= \underset{\control}{\mathrm{argmin}}\; J(\control,\state) \\
    \mathrm{s.t.}\quad
    \dot h_k(\state,\control)
    &\ge -\alpha\!\left(h_k(\state)\right),\quad \forall k\in\{1,\dots,K\}, \\
    \control_{lb} \le \control &\le \control_{ub}.
\end{aligned}
\end{equation}
Since the system is control-affine, each CBF constraint is affine in $\boldsymbol{u}$, more specifically
\begin{equation}
\label{eq:method:affine_hdot}
\dot h_k(\state,\control) = c_k(\state) + \boldsymbol{d}_k(\state)^\top \control.
\end{equation}
We elect to use the standard quadratic objective that penalizes deviation from the nominal control input $\control_{\mathrm{ref}}(\state)$
\begin{equation}
\label{eq:method:cost}
J(\boldsymbol{u},\boldsymbol{x}) =
(\boldsymbol{u}-\boldsymbol{u}_{\mathrm{ref}}(\boldsymbol{x}))^\top \mathbf{P}\,
(\boldsymbol{u}-\boldsymbol{u}_{\mathrm{ref}}(\boldsymbol{x})),
\end{equation}
with $\mathbf{P}\succ 0$. Since $J$ is quadratic and constraints are linear in $\control$, optimization (\ref{eq:method:optimization_cbfqp}) is a Quadratic Program (QP), solved online to obtain $\control^*$.
Overall, the online operation of the proposed OGM-CBF-QP follows Algorithm \ref{alg:method:deploy}.

\begin{algorithm}[htbp]
\caption{Map-based Safe Control Algorithm}
\label{alg:method:deploy}
\DontPrintSemicolon
\KwInput{Observations $\observation$, position $\position$, and heading $\heading$.}
\KwOutput{Control input $\control$.}
\While{the robot is running}
 {    
    $\observation \leftarrow O_1(.)$ , $(\position,\heading) \leftarrow O_2(.)$ \Comment{\footnotesize{Gather observations.}}\\
    $\gridmap \leftarrow \text{update}(\gridmap, \position, \heading, \observation)$ \Comment{\footnotesize{Update OGM, \eqref{eq:background:ogm}.}}\\
    $\stateSpaceSafe \leftarrow \text{process}\ \gridmap$ \Comment{\footnotesize{Compute safe set, \eqref{eq:background:unsafe}.}}\\
    \For(\Comment{\footnotesize{For each pyramid level:}}){$k \in \{1,\dots,K\}$}  
    {     
    $\sdf_k(.), \nabla_{\position} \sdf_k(.)$ \Comment{\footnotesize{compute SDF, \cref{sc.sdf_pyr};}}\\
    $\sdfSmooth_k(.), \nabla_{\position} \sdfSmooth_k(.)$ \Comment{\footnotesize{interpolate SDF, \cref{sec:method:smooth};}}\\
    $\cbf_k(\state), c_k(\state), d_k(\state)$ \Comment{\footnotesize{compute CBF, \eqref{eq:method:cbf} and \eqref{eq:method:affine_hdot}.}}\\
    }
    $\control^\star \leftarrow \text{optimize QP}$ \Comment{\footnotesize{Solve the OGM-CBF-QP \eqref{eq:method:optimization_cbfqp}.}}\\
    $\text{apply}\ \control^\star$
}
\end{algorithm}


\section{Experimental Evaluation}\label{sec:results}
To experimentally verify the proposed OGM-CBF, a quantitative and qualitative evaluation first compares the proposed approach against memory-unaware baselines.
In the rest of the presented results, we investigate the effect of the SDF pyramid and the resulting multi-level formulation, and report on deployments on a small warehouse robot and an Articulated Frame Steering (AFS) autonomous wheel loader.

\paragraph*{A Note on Common Kinematics and  Control Objectives}\label{sec:common_ctrl}
In~\cref{sc.effect_of_memory,sc.effect_multi_level,sc.mir}, the robots employ a common system with the state ${\boldsymbol{x}= [x,y, \psi]^\top}$ (the coordinates of the vehicle's center in the $2$D plane, and the heading), with $\boldsymbol{u}=[v,w]^T$ (linear and angular speeds) as the control inputs.
Safe navigation is deployed by Algorithm \ref{alg:method:deploy} and the quadratic optimization framework (\ref{eq:method:optimization_cbfqp}), with the nominal control objective to maintain a target heading per the cost $J(\control,\state)
=\frac12\bigl(v-v_{\mathrm{ref}}\bigr)^2
+\frac12\bigl(\omega-\omega_{\mathrm{ref}}\bigr)^2$ with reference velocities $v_\mathrm{ref}$ and $\omega_\mathrm{ref}= -k_{\psi}(\psi-\psi_\mathrm{ref})$.
The kinematics of the AFS wheel loader in~\cref{sc.wheelloader} differ, and thus are described therein.

\subsection{Obstacle Memory in Limited FoV Operation}
\label{sc.effect_of_memory}
Two memory-unaware baselines are considered for a quantitative and qualitative comparison in the CARLA autonomous driving simulator~\cite{dosovitskiy2017carla}. First, the SDF-based variant of V-CBF~\cite{abdi2023safe} is selected to represent memory-unaware methods due to its native support for depth cameras, which exemplify limited-FoV operation. Second, a memory-unaware modification of the OGM-CBF (MU-OGM-CBF) employs the current sensory input only to construct non-persistent grid maps, used for an ablation of the memory/map persistency.

Both the proposed approach and the baselines are provided with the same forward-facing \SI{60}{\degree} FoV depth camera and the vehicle odometry. 
The depth imagery is used natively in the V-CBF, while the OGM-CBF variants build an EKF-based elevation grid map~\cite{bayer_cache_2020} and refine it into an occupancy grid map with $\cellsize = \SI{0.1}{\meter}$ cells.
Steering and throttle commands are calculated using the standard kinematic bicycle model in a $2$D workspace~\cite{rajamani2011vehicle}, and the vehicle tracks the nominal target heading $\psi_{\mathrm{ref}}$ while guaranteeing safety in a road scenario with static obstacles. Using the nominal common heading-tracking controller, we set $v_{\mathrm{ref}}=\SI{3}{\meter\per\second}$ and the heading gain $k_\psi=0.5$.
During all runs, the (MU)-OGM-CBF were configured with $\alpha(\cbf){=}0.6\cbf$, $v\!\in[0,3]\,\si{\meter\per\second}$, $\omega\!\in[-4\pi,4\pi]\,\si{\radian\per\second}$, $l_a{=}0.25$, $l_s{=}-0.25$, obstacle inflation radius $\SI{2.1}{\meter}$, and $K=3$ levels of pyramid with $\sigma=1$ Gaussian kernel; and V-CBF with $\alpha(\cbf){=}0.7\cbf$, and the same bounds on $v,\omega$.

\begin{figure}[!htb]
  \scalebox{-1}[1]{\includegraphics[width=0.49\columnwidth,trim={0 12cm 0 3cm},clip]{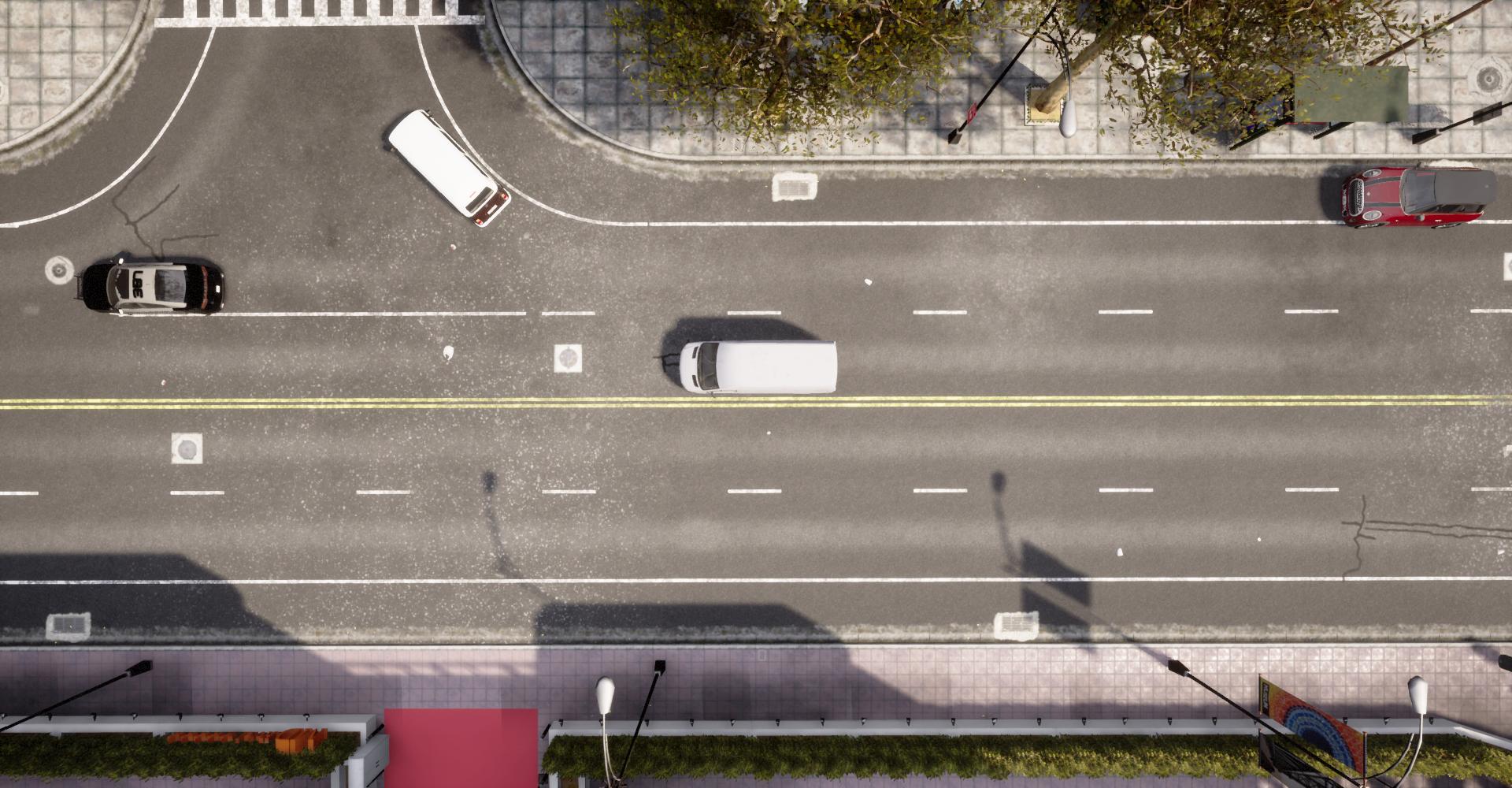}}
  \scalebox{-1}[1]{\includegraphics[width=0.49\columnwidth,trim={0 12cm 0 3cm},clip]{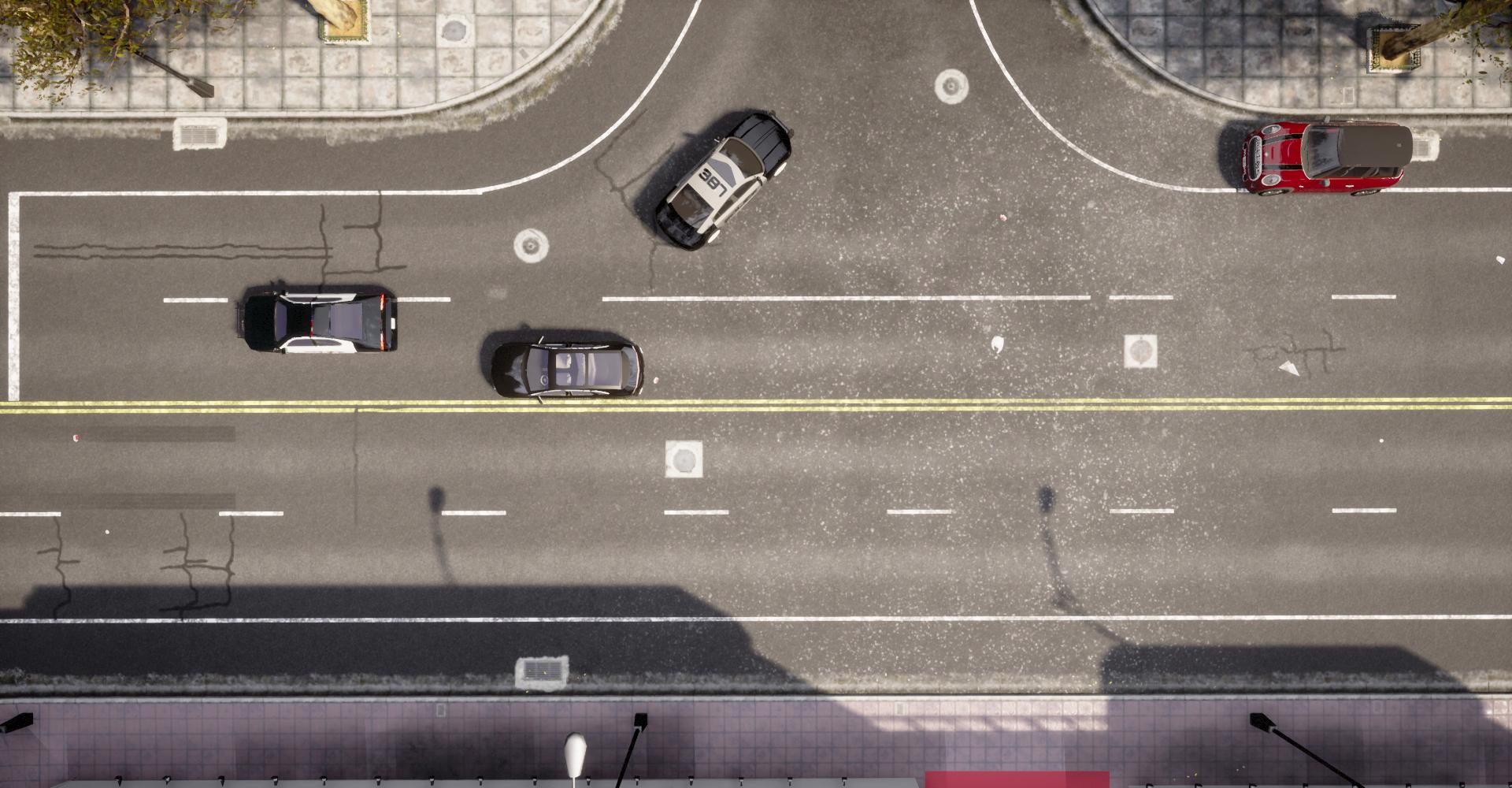}}
  \caption{\label{fig:results:carla_sc}%
  Overhead view of the (left) \texttt{Open} and (right) \texttt{Trap} scenarios used in the quantitative evaluation.
  Example starting position of the ego vehicle in red, in the left of each view.
  }
\end{figure}
\begin{table}[htb]
  \centering
\caption{%
  Simulated road scenarios, collision avoidance ($\%$ of $19$ runs) and average minimum distance to obstacle (center to center) ($\mu\pm\sigma$ of successful runs). 
  \label{tbl:res:carla}
  }
  \begin{tabular}{l rr rr}
\toprule
                          & \multicolumn{2}{c}{\texttt{                     Open}} & \multicolumn{2}{c}{\texttt{                     Trap}} \\
                    Model & Succ. & $L2_\text{min}$ $(\mu\pm\sigma)$ & Succ. & $L2_\text{min}$ $(\mu\pm\sigma)$ \\
\midrule
         \textbf{OGM-CBF} & $\SI{100}{\percent}$ & $\num{5.53} \pm \SI{0.55}{\meter}$ & $\SI{100}{\percent}$ & $\num{5.18} \pm \SI{0.48}{\meter}$ \\
               MU-OGM-CBF & $\SI{ 84}{\percent}$ & $\num{2.75} \pm \SI{0.47}{\meter}$ & $\SI{ 37}{\percent}$ & $\num{2.95} \pm \SI{0.41}{\meter}$ \\
                    V-CBF & $\SI{ 53}{\percent}$ & $\num{2.81} \pm \SI{0.28}{\meter}$ & $\SI{ 37}{\percent}$ & $\num{2.99} \pm \SI{0.54}{\meter}$ \\
\midrule
\end{tabular}
  \vspace{-1em}
\end{table}
The quantitative evaluation is carried out in the two scenarios illustrated in \cref{fig:results:carla_sc}.
\begin{figure}[!htb]
  \scalebox{-1}[1]{\includegraphics[width=\columnwidth,trim={0 3cm 0 3cm},clip]{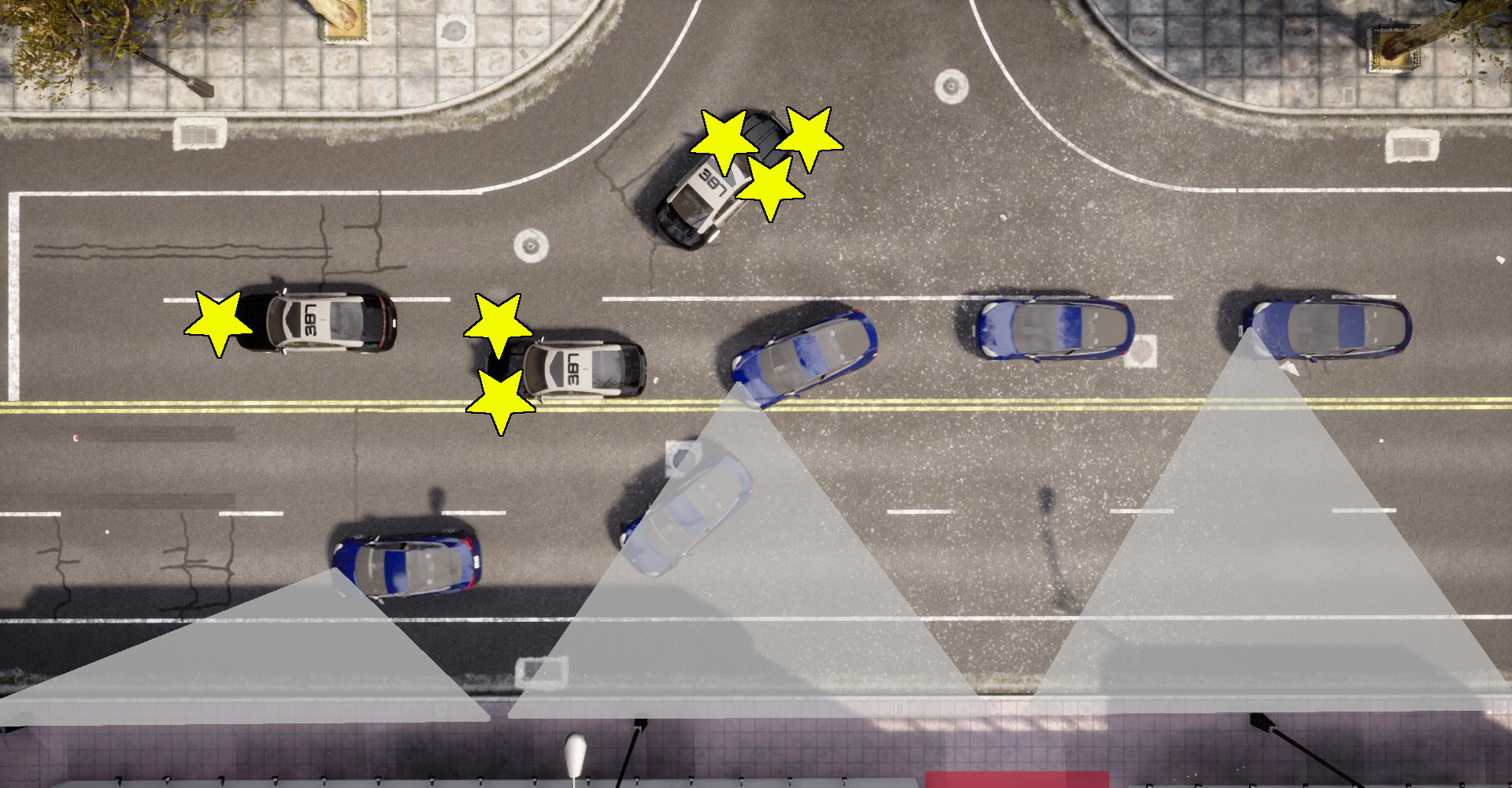}}\\[0.1em]
  \phantom{}
  \hfill
  \includegraphics[height=0.65cm]{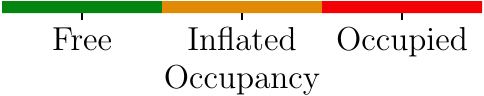}
  \hfill
  \phantom{}\\[0em]
  \scalebox{-1}[1]{\includegraphics[width=0.3\columnwidth]{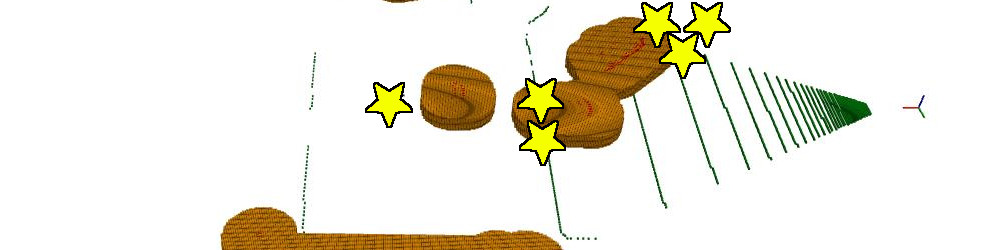}}
  \hfill
  \scalebox{-1}[1]{\includegraphics[width=0.3\columnwidth]{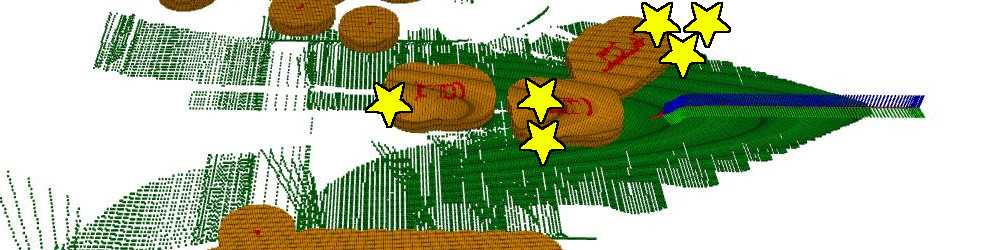}}
  \hfill
  \scalebox{-1}[1]{\includegraphics[width=0.3\columnwidth]{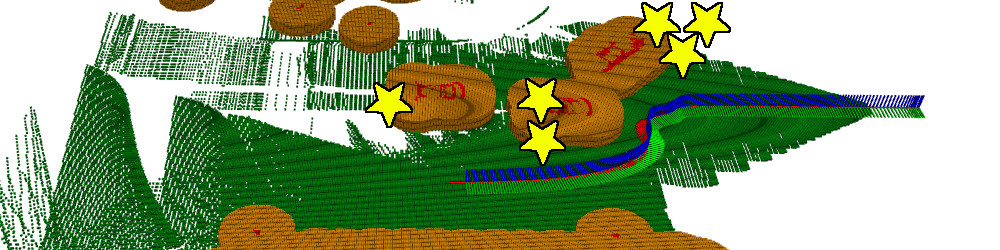}}\\[-1.3em]
  \subfloat[\textbf{Proposed OGM-CBF} builds persistent environment representation by aggregating the limited FoV scans.]{\phantom{\hspace \columnwidth}}\\[0.5em]
  \scalebox{-1}[1]{\includegraphics[width=\columnwidth,trim={0 12cm 0 3cm},clip]{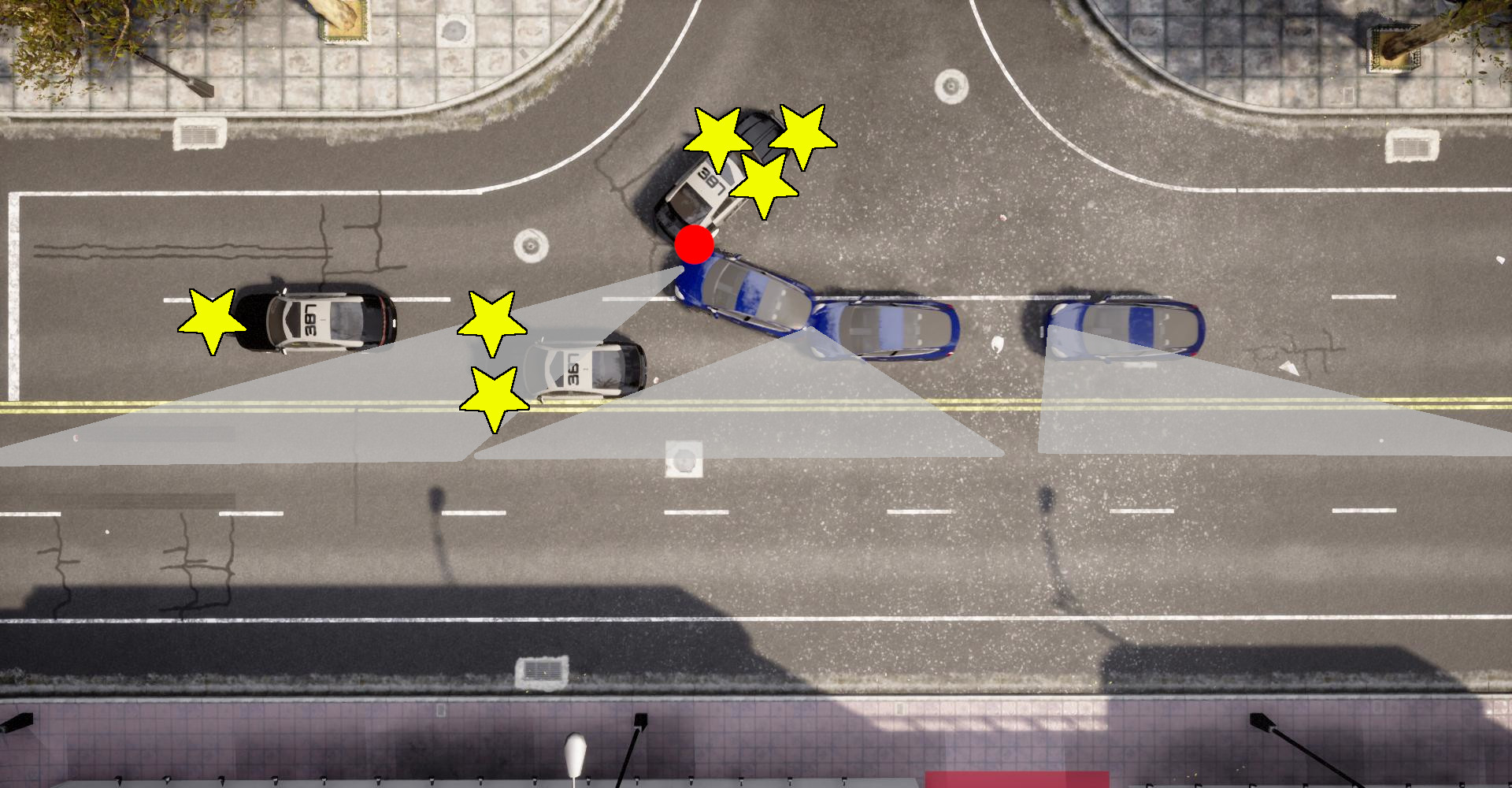}}\\[0.1em]
  \phantom{}
  \hfill
  \includegraphics[height=0.65cm]{fig/colorbar_trav-crop.pdf}
  \hfill
  \phantom{}\\[0em]
  \scalebox{-1}[1]{\includegraphics[width=0.32\columnwidth]{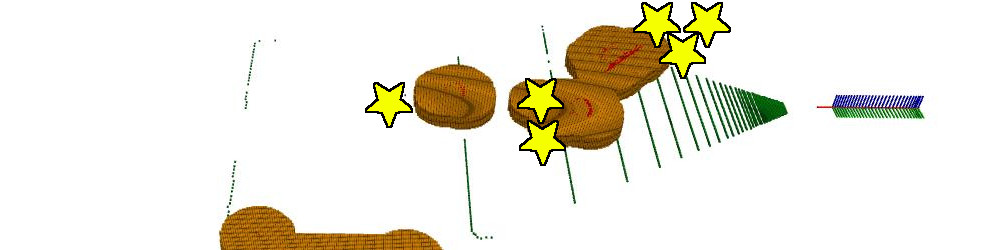}}
  \hfill
  \scalebox{-1}[1]{\includegraphics[width=0.32\columnwidth]{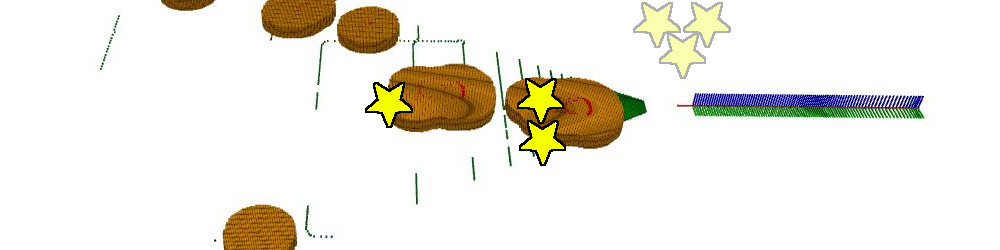}}
  \hfill
  \scalebox{-1}[1]{\includegraphics[width=0.32\columnwidth]{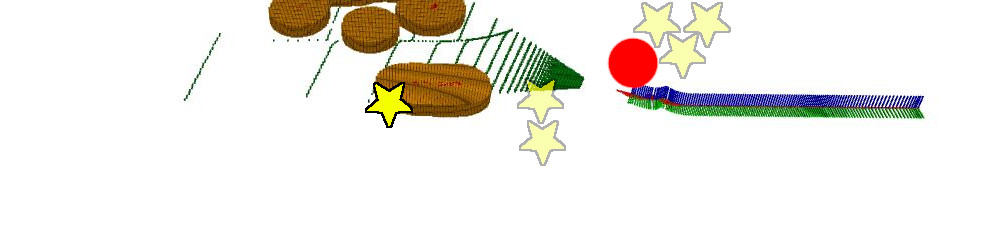}}\\[-1.3em]
  \subfloat[Memory-unaware MU-OGM-CBF uses the same representation, but has no persistency and considers only the current, limited FoV.]{\phantom{\hspace \columnwidth}}\\[0.5em]
  \scalebox{-1}[1]{\includegraphics[width=\columnwidth,trim={0 12cm 0 3cm},clip]{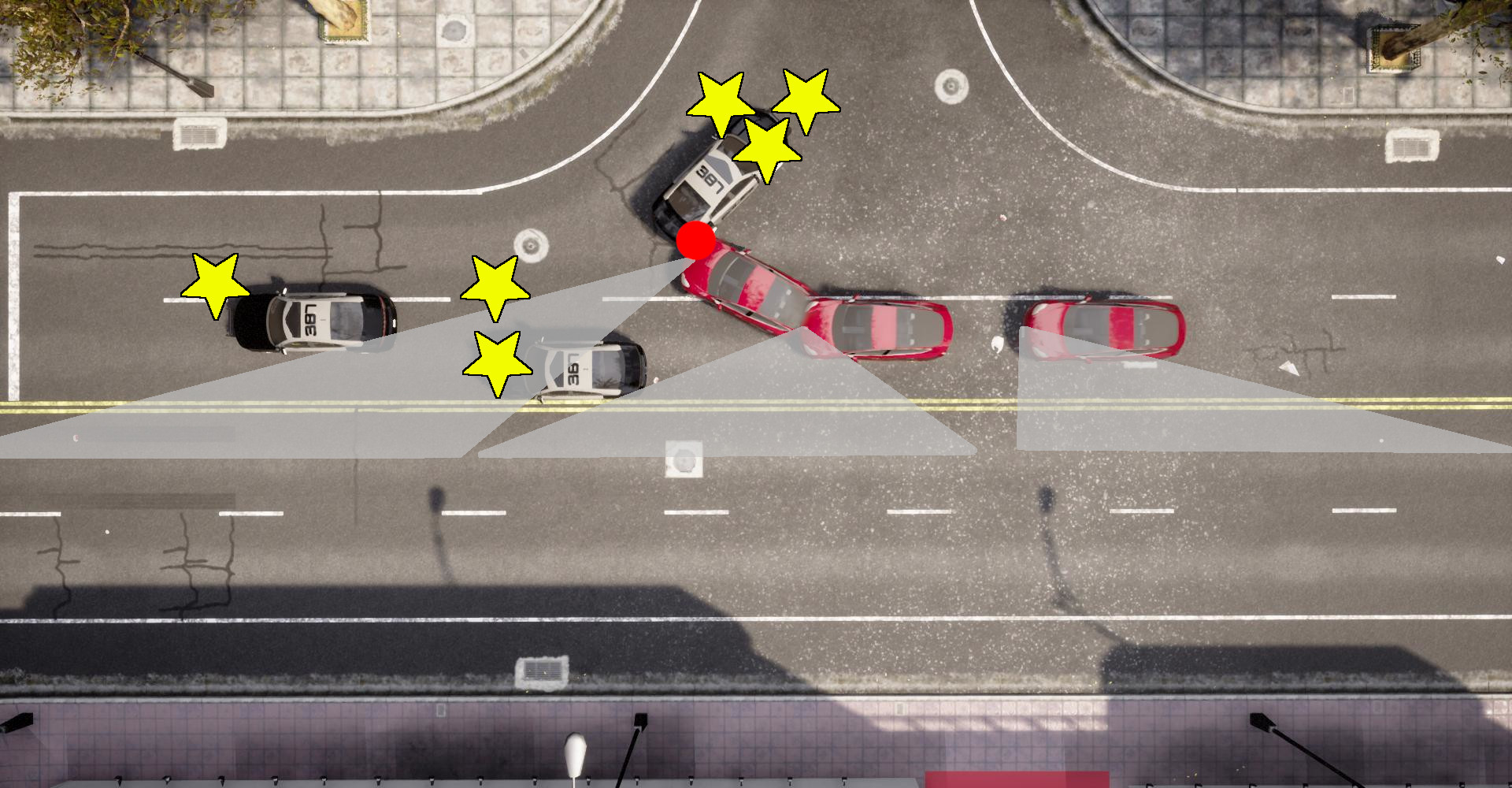}}\\[0.5em]
  \scalebox{-1}[1]{\includegraphics[height=1.35cm]{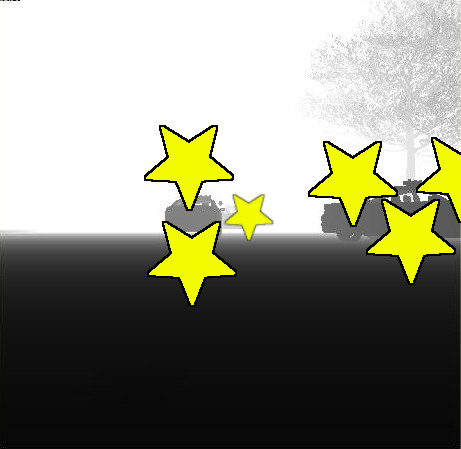}}
  \hfill
  \scalebox{-1}[1]{\includegraphics[height=1.35cm]{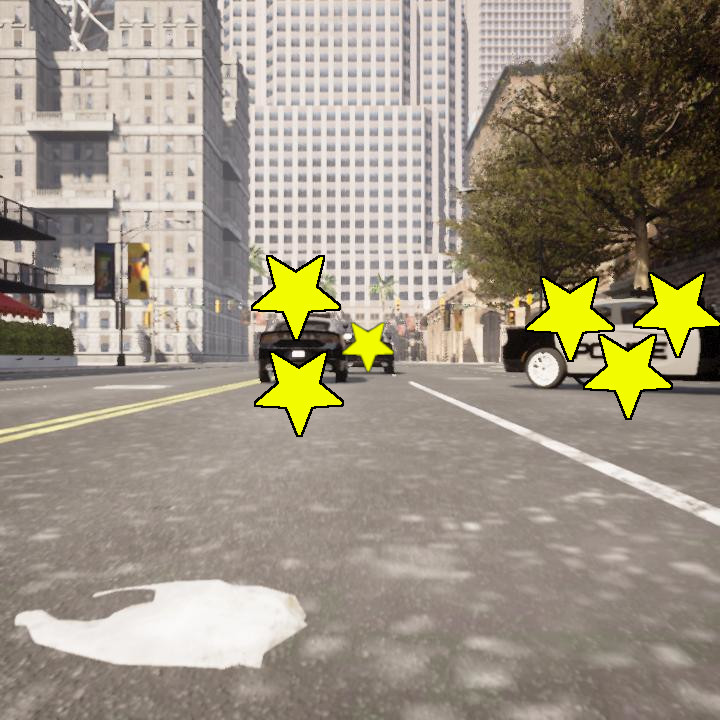}}
  \hfill
  \scalebox{-1}[1]{\includegraphics[height=1.35cm]{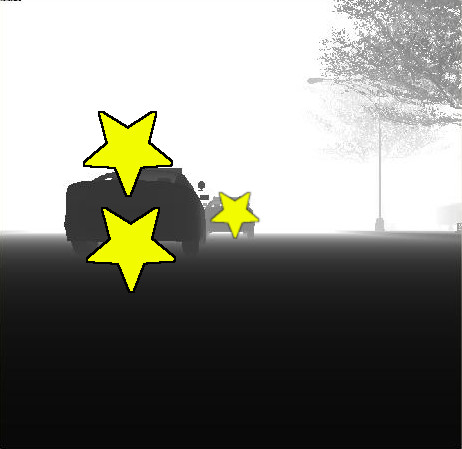}}
  \hfill
  \scalebox{-1}[1]{\includegraphics[height=1.35cm]{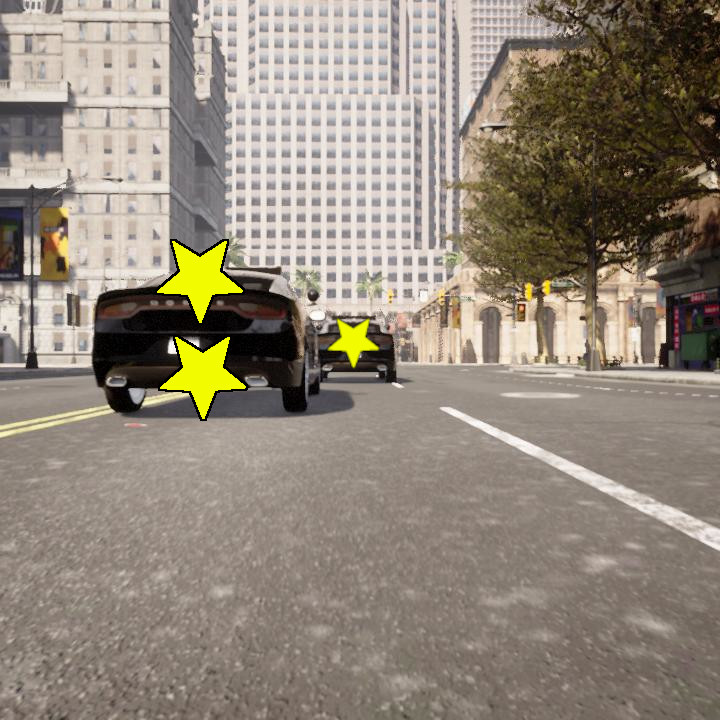}}
  \hfill
  \scalebox{-1}[1]{\includegraphics[height=1.35cm]{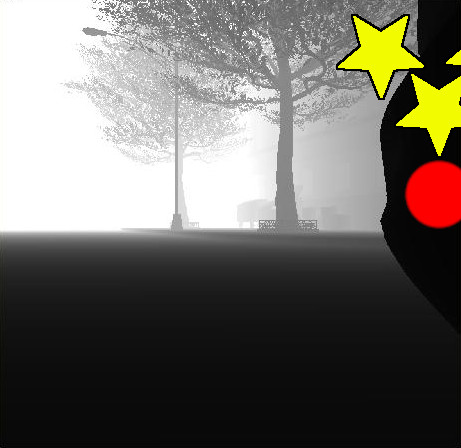}}
  \hfill
  \scalebox{-1}[1]{\includegraphics[height=1.35cm]{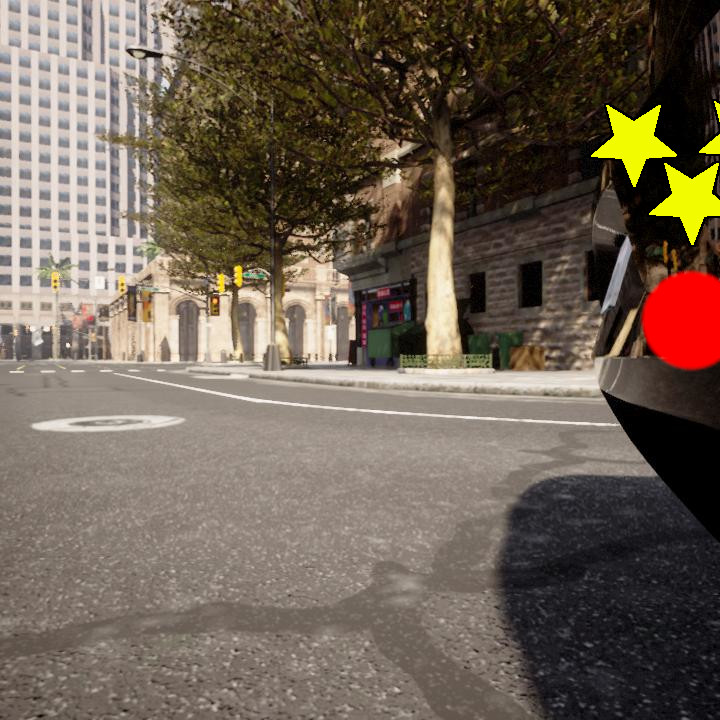}}\\[-1.3em]
  \subfloat[Memory-unaware V-CBF considers only the current limited FoV.]{\phantom{\hspace \columnwidth}}
  \caption{\label{fig:results:carla}%
  Proposed OGM-CBF compared to baselines without memory of out-of-FoV obstacles.
  (a) A vehicle controlled using the proposed OGM-CBF and three snapshots of its occupancy grid map built online during the operation.
  Note, while the grid map is $2$D, the inflated obstacles are above the ground level for visual clarity.
  The proposed OGM-CBF retains the knowledge of obstacles after they leave the vehicle's FoV, and therefore avoids all the obstacles.
  (b) The same view for MU-OGM-CBF without grid map memory.
  The occupancy grid map snapshots are thus individual scans processed into one-shot grid maps.  
  (c) A simulated car controlled using the V-CBF without grid map memory and three snapshots of its depth perception with RGB reference. Both memory-unaware baselines lose awareness of the top-located (marked by three stars) obstacle while navigating, and steer towards it when avoiding the bottom-located (marked by two stars) in-FoV obstacle, resulting in a collision (red marker).
  }
\end{figure}
\cref{tbl:res:carla} overviews the behaviors in the two scenarios, each with $19$ uniformly distributed starting positions. 
In the \texttt{Open} scenario, the vehicle faces spaced obstacles.
While the proposed approach navigates the scenario safely with \SI{100}{\percent} success rate, the reactive, memory-unaware baselines suffer from their limited FoV.
In particular, the V-CBF succeeds in \SI{53}{\percent} runs, while the MU-OGM-CBF has \SI{84}{\percent} success rate. Intuitively, the MU-OGM-CBF benefits from the obstacle inflation in the Euclidean OGM space, whereas the V-CBF relies solely on projected image-frame perception.
The proposed OGM-CBF combines this advantage with persistent obstacle memory, achieving the highest success rate. Moreover, OGM-CBF exhibits higher average minimum distance to obstacle in the successful runs, indicating wider avoidance and thus better safety, albeit at the cost of losing the inherent efficiency of tight avoidance.

The obstacles in the \texttt{Trap} scenario are tighter, and placed so that a vehicle relying only on its FoV may be prompted to steer away from one obstacle into another.  
The proposed approach leverages the grid map to avoid all obstacles, while both the MU-OGM-CBF and V-CBF succeed only in \SI{37}{\percent} of the starting configurations.
Again, the proposed approach exhibits larger average minimum distance to obstacle, suggesting wider avoidance.
Moreover, \cref{fig:results:carla} further visualizes how the memory-unaware methods enter states where avoidance of one obstacle steers them into another despite it being within the FoV at a previous time step.

\subsection{Effect of Multi-level CBF Formulation}
\label{sc.effect_multi_level}
\begin{figure}[!htb]
  \includegraphics[width=\columnwidth]{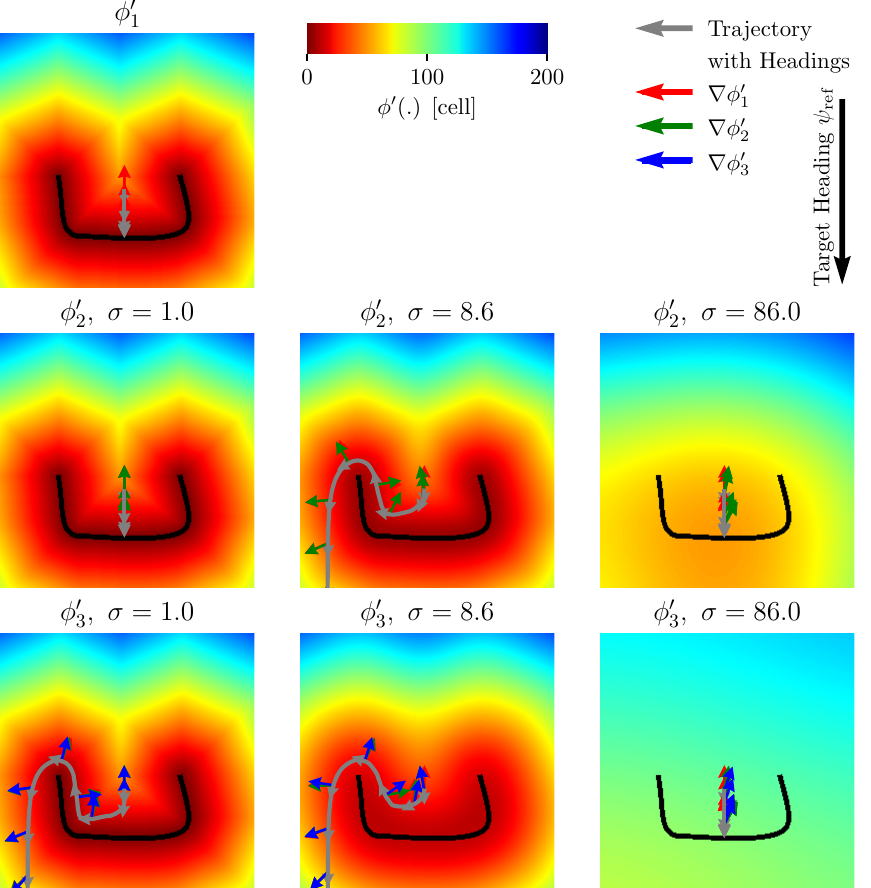}
\caption{\label{fig:results:pyramid}%
  The effect of the pyramid SDF transform when facing a heading-orthogonal obstacle (black).
  Pyramid levels $k=1$ (baseline SDF), $2$, and $3$ on the vertical axis; Gaussian blur applied before each pyramid transform on the horizontal axis. 
  (top left) The baseline SDF gradients $\nabla\sdfSmooth_{1}$ are directly opposite to the heading, and thus without the additional pyramid levels the vehicle lacks the steering authority for safe behaviors other than stopping in front of the obstacle.
  (bottom left, center) The transformed SDF gradients $\nabla\sdfSmooth_{k},k=2,3$ differ from the baseline gradients, and thus the vehicle is provided with the steering authority to navigate around the obstacle.
  (center) Applying stronger Gaussian kernels before the pyramid transforms shifts the gradients further away from the baseline degenerate gradient, (right) but must be tuned to the grid size to avoid loss of information and steering authority.
  }
\end{figure}
  A $2$D differential drive kinematic simulator is employed to investigate the effects of the SDF pyramid transform when navigating obstacle surfaces locally orthogonal to robot heading.
\cref{fig:results:pyramid} highlights the SDF gradient of the individual levels, and the resulting safe avoidance trajectory in a U-shaped obstacle scenario, given the desired target heading $\psi_{\mathrm{ref}}$. 
The baseline constructs OGM-CBF only w.r.t. the level $1$ SDF (no pyramid transform).
As seen, for a heading-orthogonal obstacle, the SDF gradient $\nabla_{\position} \sdfSmooth_k (.), k=1$ is directly opposite to the heading, demonstrating the gradient degeneracy described in~\cref{sc.sdf_pyr}.
Therefore, while the system stays in the safe set, stopping in front of the obstacle, the baseline is immobilized since it lacks steering authority.
On the other hand, the proposed method employs a three-level pyramid ($2$ pyramid transforms applied to the SDF of the OGM).
In particular, the gradients of SDFs on levels $2$ and $3$ (($\nabla_{\position} \sdfSmooth_k (.), k=2,3$)) are not directly opposite to the heading, injecting the authority to navigate around the obstacle, see~\cref{fig:results:pyramid}.
However, it should be noted that there is only a particular subspace on the shape of the Gaussian kernel $\mathcal{G}_\sigma$ and number of pyramid levels where one can prevent loss of information from high blur and keep efficacy. Besides, consider that Gaussian blurring should not be applied directly to the $h_0$ as it smooths $h_0$ around $h_0=0$, shifting the zero-superlevel set $\superlevel$ of the primary CBF.

\subsection{Deployment on a Warehouse Robot}
\label{sc.mir}
\begin{figure}[!htb]
  \phantom{}
  \hspace{0.31\columnwidth}
  \hfill
  \phantom{}
  \includegraphics[height=0.5cm]{fig/colorbar_trav-crop.pdf}
  \hfill
  \includegraphics[height=0.5cm]{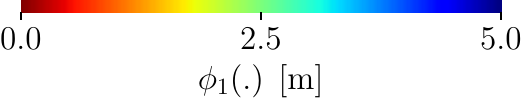}
  \phantom{}\\[0.2em]
  \includegraphics[height=2.7cm]{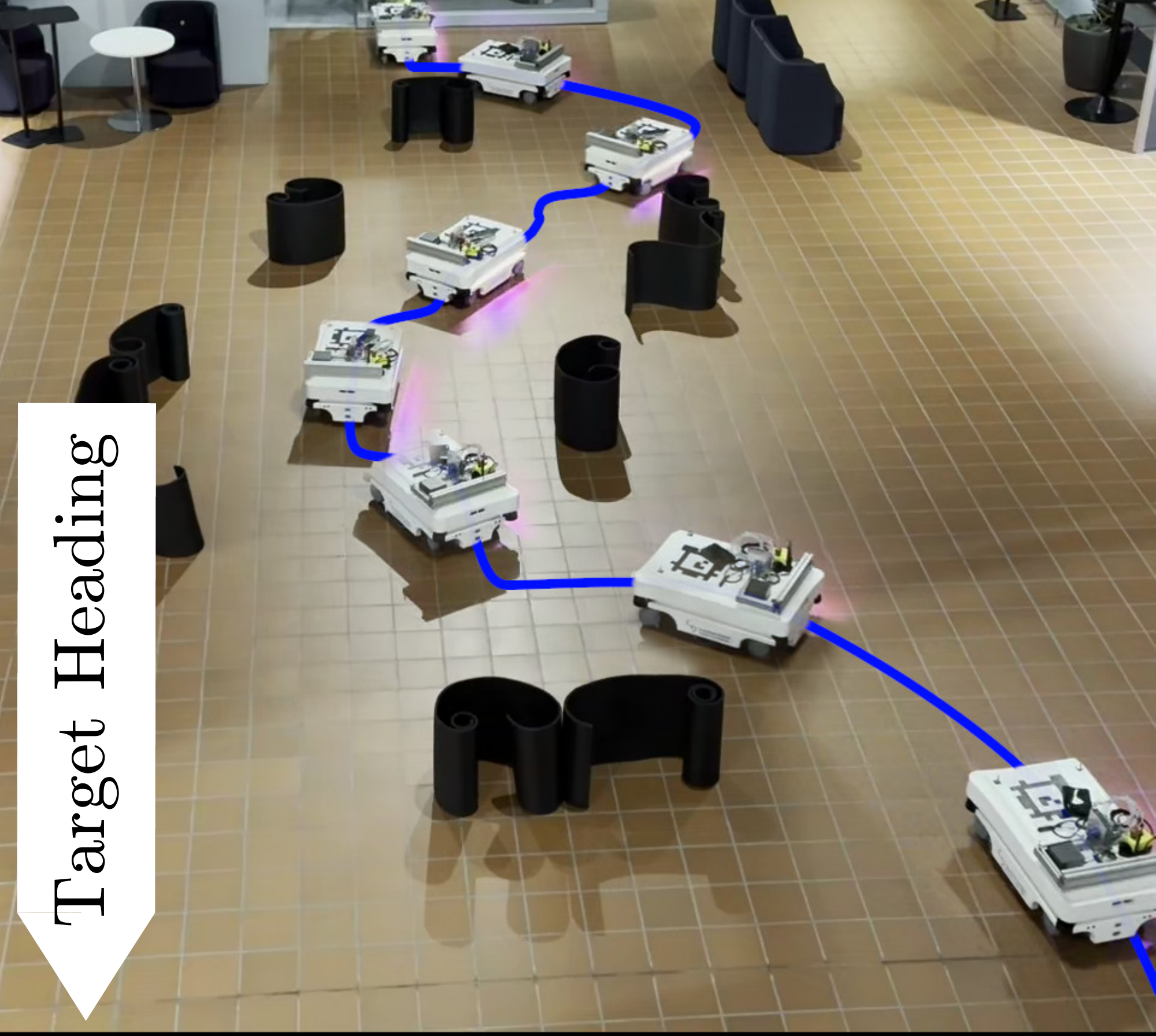}
  \hfill
  \includegraphics[height=2.7cm]{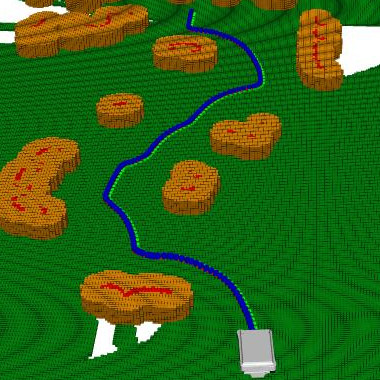}
  \hfill
  \includegraphics[height=2.7cm]{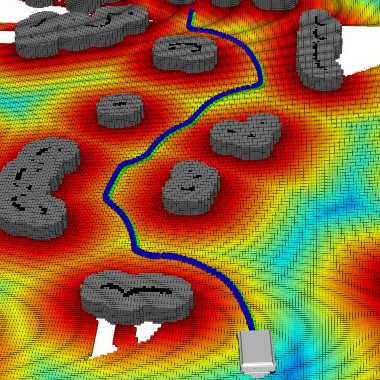}
  \phantom{}\\[0.5em]
  \includegraphics[width=0.97\columnwidth]{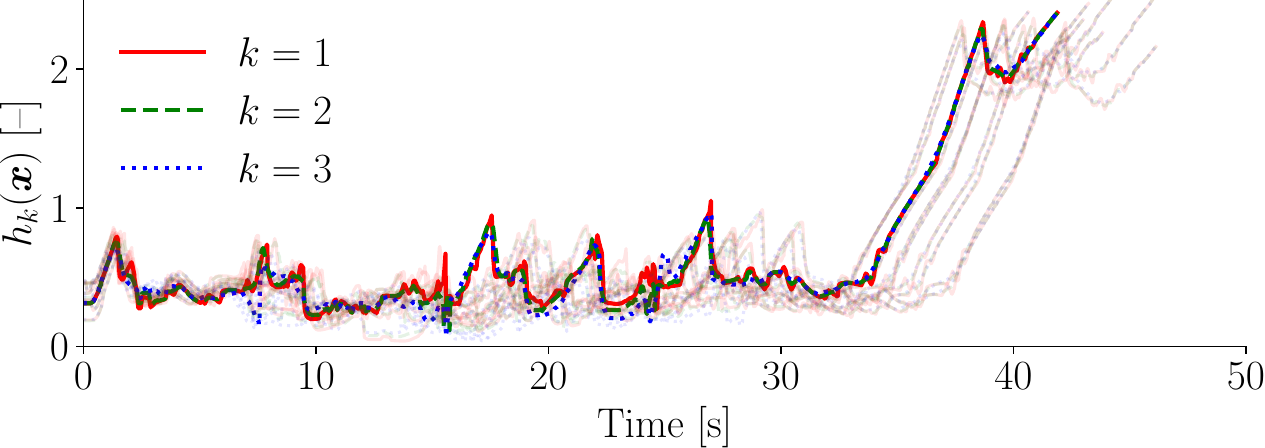}
  \hfill
  \phantom{}\\[-1.5em]
  \subfloat[Clutter scenario.]{\phantom{\hspace \columnwidth}}\\[1em]
  \phantom{}
  \hspace{0.31\columnwidth}
  \hfill
  \phantom{}
  \includegraphics[height=0.5cm]{fig/colorbar_trav-crop.pdf}
  \hfill
  \includegraphics[height=0.5cm]{fig/colorbar_sdf_5-crop.pdf}
  \phantom{}\\[0.2em]
  \includegraphics[height=2.7cm]{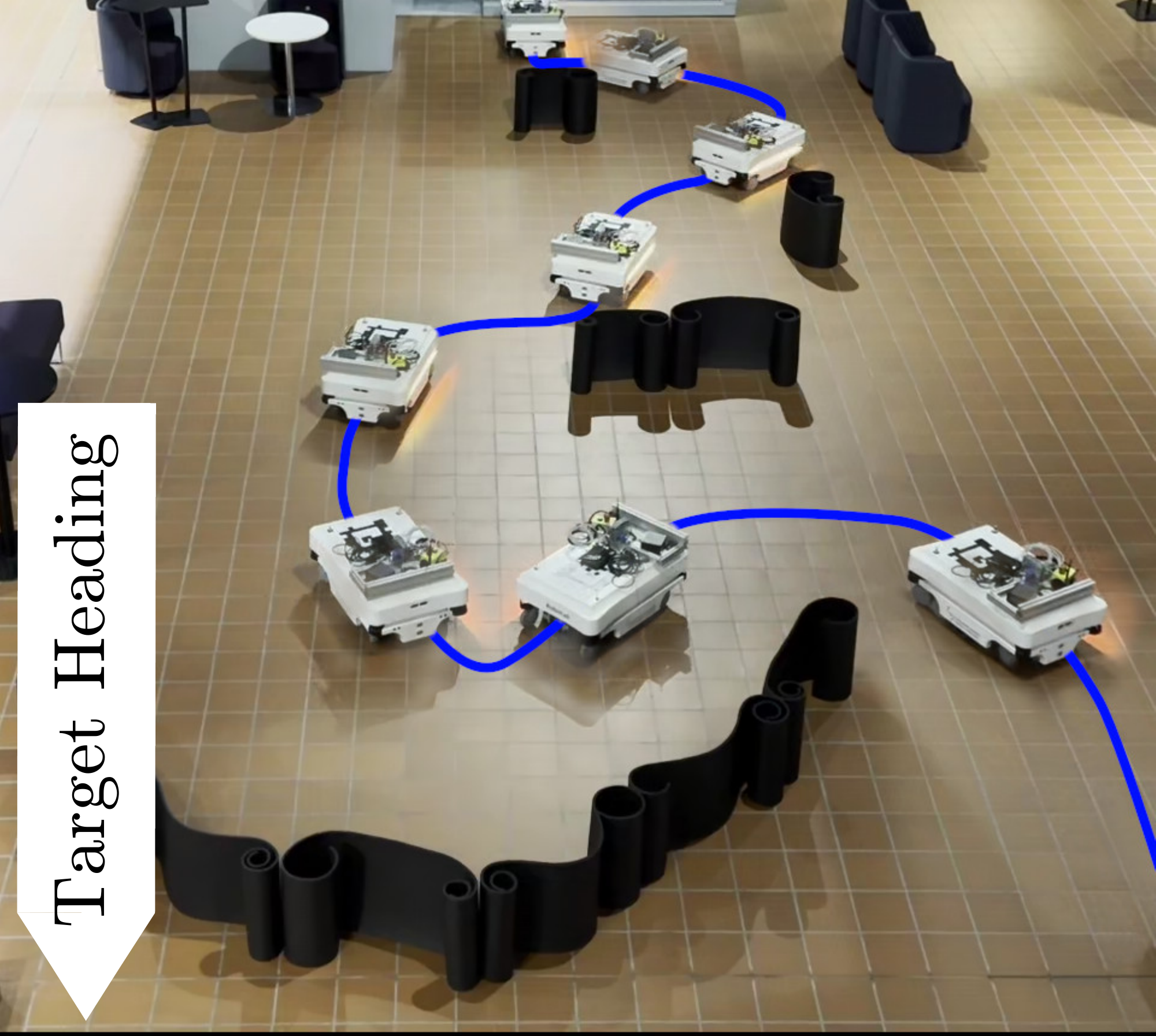}
  \hfill
  \includegraphics[height=2.7cm]{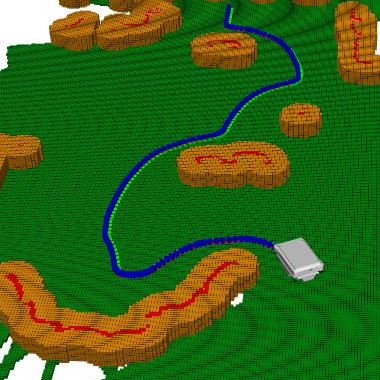}
  \hfill
  \includegraphics[height=2.7cm]{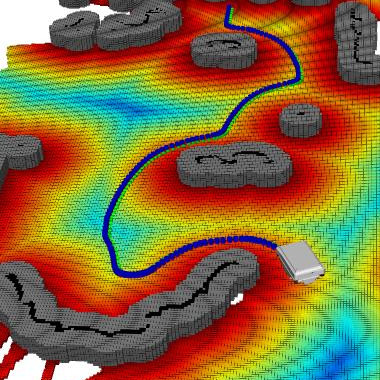}
  \phantom{}\\[0.5em]
  \includegraphics[width=0.97\columnwidth]{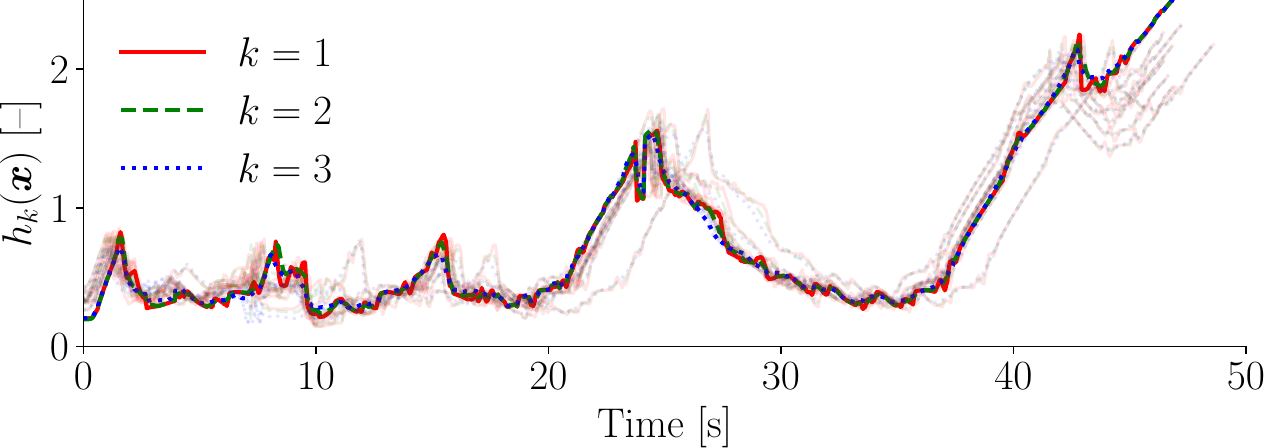}
  \hfill
  \phantom{}\\[-1.5em]
\subfloat[Concave scenario.]{\phantom{\hspace \columnwidth}}
\caption{%
  Navigation of the \texttt{MiR100} warehouse robot using the proposed OGM-CBF in the (a) clutter and (b) concave scenarios.
  In both, the robot is tasked to drive in the direction of the bottom image edge.
  (top left) Overhead view of the scenarios with the robot shown every \SI{5}{\second}, (top mid) the occupancy grid map at the end of the respective scenario with overlayed robot path, (top right) and the respective SDF.    
  Note, while the grid map is $2$D, the inflated obstacles are above the ground level for visual clarity.
  (bottom) The development of the CBF while navigating the respective scenario.
  The CBFs $\cbf_{k}(\state)$ corresponding to each SDF pyramid level $k$ are shown separately.
  In each scenario, one representative run is in bold, and the remaining runs are muted for reference. 
  }
  \vspace{-1em}
\label{fig:results:mir_vis}
\end{figure}

The proposed approach has been deployed on \texttt{MiR100}, a warehouse differential drive platform with a front-left \texttt{SICK S300} $\SI{270}{\degree}$-FOV $2D$ laser scanner and integrated fusion odometry. The two scenarios in~\cref{fig:results:mir_vis} are each repeated $5$ times in $2$ sensing configurations for $20$ runs in total. The scenarios model small obstacle clutter and a large concave obstacle, respectively, both using short and long range sensing, with the $2D$ scanner limited to $\SI{5}{\meter}$ and $\SI{25}{\meter}$, respectively.

During the experiments, the system is initialized with no prior map of the environment, and builds the occupancy grid map with $\cellsize = \SI{0.05}{\meter}$ cells online during the navigation;
the robot embodiment is represented by $\SI{0.35}{\meter}$ obstacle inflation.
The utilized safe control synthesis described in (\ref{eq:method:optimization_cbfqp}) is configured with $\alpha(\cbf)=0.3\cbf$, $l_a=-l_s=0.25$, control bounds $v\!\in[-0.5,0.5]\,\si{\meter\per\second}$, $\omega\!\in[-\frac\pi4,\frac\pi4]\,\si{\radian\per\second}$, and $K=3$ levels of pyramid with $\sigma=1$ Gaussian kernel.
The nominal heading-tracker controller uses $v_{\mathrm{ref}}=\SI{0.5}{\meter\per\second}$ and the heading gain $k_{\psi}=0.5$.

During the deployment, the system has successfully navigated both scenarios, following similar paths over the repeated runs, with the perception, mapping, and OGM-CBF-QP pipeline running  at $\SI{10}{Hz}$ on an NVIDIA Jetson AGX Orin. Besides, the evolution of the CBF values over the individual runs indicates that the proposed system successfully and safely repeatedly avoids obstacles in both scenarios, see~\cref{fig:results:mir_vis}.    

\subsection{Deployment on an Articulated Wheel Loader}\label{sc.wheelloader}
The proposed approach has been deployed on a $5\,\mathrm{ton}$ AFS wheel loader. As described in \cite{toulkani2025safety},\cite{karki2025reactive}, the AFS front body kinematic model is
\begin{equation}
\begin{aligned}
\dot x_f &= v_f\cos\theta_f, & \quad \dot y_f &= v_f\sin\theta_f, \\
\dot\theta_f &= \frac{v_f\sin\beta + l_r\dot\beta}{l_f\cos\beta + l_r},  & \quad \dot\beta &= u_{\dot\beta},
\end{aligned}
\end{equation}
with state $\boldsymbol{x}=[x_f,\,y_f,\,\theta_f,\,\beta]^\top$ and control $\boldsymbol{u}=[v_f,\,u_{\dot\beta}]^\top$, where $(x_f,y_f)$ is the front-body reference-point position,
$\theta_f$ the front-body heading,
$\beta$ the articulation (hinge) angle between front and rear bodies,
$l_f,l_r$ are the joint-to-front and -rear axle distances, respectively,
$v_f$ is the front-body forward speed,
and $u_{\dot\beta}$ is the commanded articulation rate (i.e., $\dot\beta=u_{\dot\beta}$).

\begin{figure}[!htb]
  \includegraphics[width=\columnwidth]{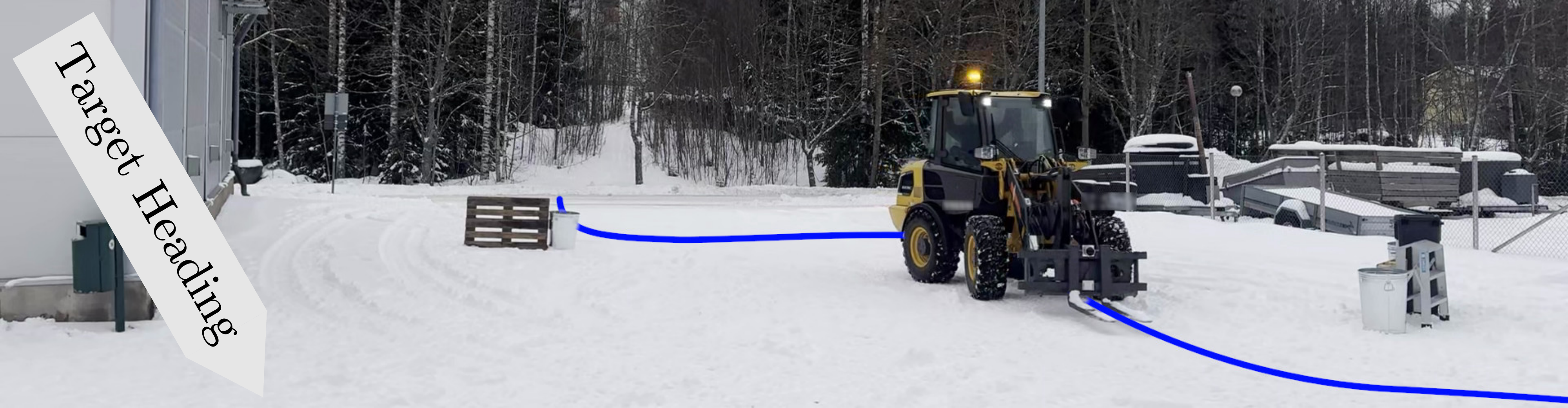}\\[0.5em]
  \phantom{}
  \hfill
  \includegraphics[height=0.65cm]{fig/colorbar_trav-crop.pdf}
  \hfill
  \phantom{}\\[0em]
  \includegraphics[width=\columnwidth]{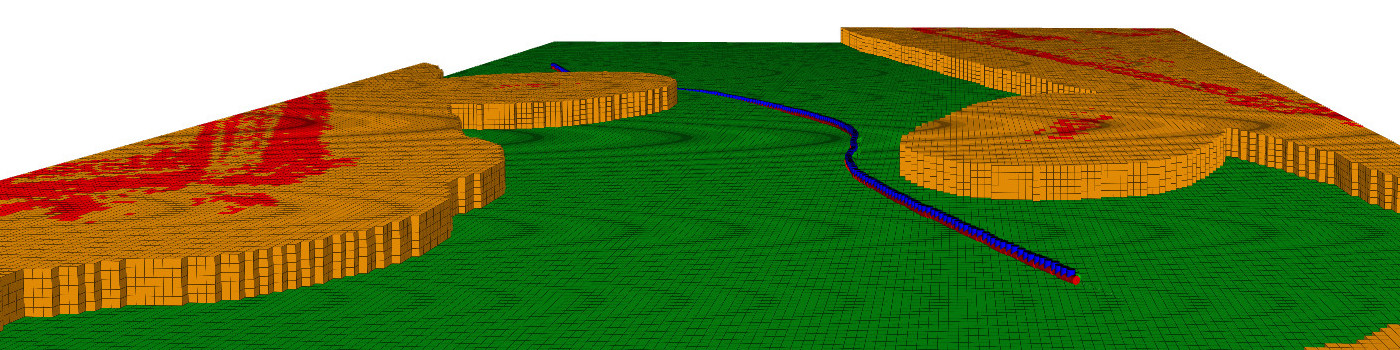}\\[.5em]
  \includegraphics[width=0.97\columnwidth]{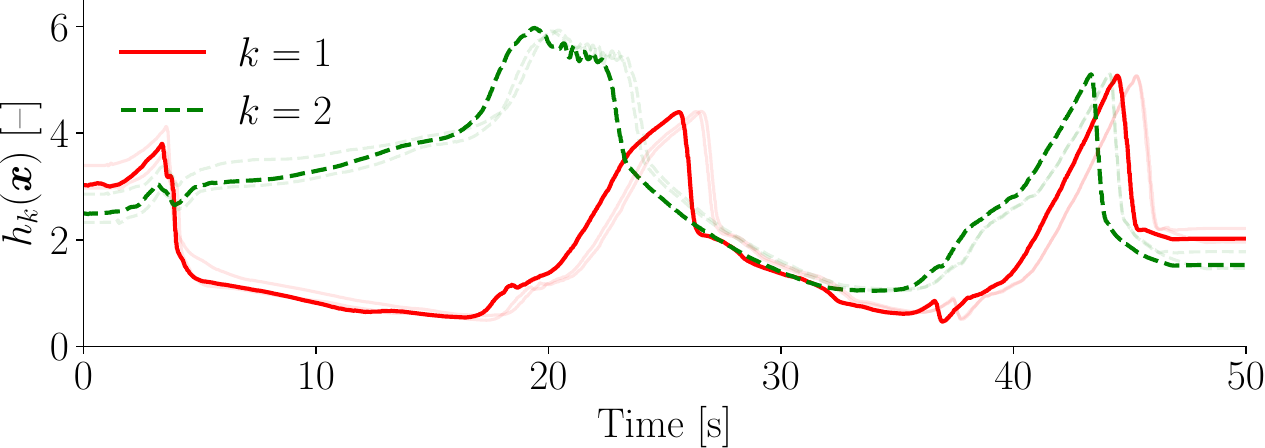}
  \hfill
  \phantom{}\\[-1.5em]
\caption{\label{fig:results:eevee}%
  Wheel loader deployment.
  (top) View of the scenario.
  (mid) The occupancy grid map at the end of the respective scenario with overlaid robot path.   
  Note, while the grid map is $2$D, the inflated obstacles are above the ground level for visual clarity.
  (bottom) The development of the CBF value $\cbf_{k}(\state)$ corresponding to each image pyramid level $k$ shown separately.
  One representative run is shown in bold, and two remaining runs are shown muted for reference. 
  }
\end{figure}
The machine is equipped with a rear-facing \texttt{Velarray M1600} LiDAR.
Hence, the machine first maps the environment using the rear-facing LiDAR, creating a memory of the obstacles in an elevation grid map that is refined into $\cellsize=\SI{0.1}{\meter}$ occupancy.
Then, the map is used to navigate in forward direction.
In particular, the machine is tasked to nominally track a target heading $\psi_{\mathrm{ref}}$ while operating safely in the static obstacle scenario presented in~\cref{fig:results:eevee}.
At each control step, the OGM-CBF-QP defined in (\ref{eq:method:optimization_cbfqp}) is solved~\cite{karki2025reactive}. The control pipeline is executed on an Intel Core i5 processor at a control frequency of \SI{100}{Hz}. Nominal motion is defined by $u_{\mathrm{ref}}=[v_{\mathrm{ref}},\,u_{\dot\beta,\mathrm{ref}}]^\top$, where $v_{\mathrm{ref}}=\SI{1.0}{\meter\per\second}$ and $\omega_{\mathrm{ref}} = -k_\psi\,(\theta_f-\psi_{\mathrm{ref}}), 
u_{\dot\beta,\mathrm{ref}} = -\frac{v_{\mathrm{ref}}}{l_r}\sin\beta
+(\frac{l_f}{l_r}\cos\beta+1)\omega_{\mathrm{ref}}.$

In the described avoidance scenario, the parameters were set to $\alpha(\cbf)=0.15\cbf$$, l_a=1.5, l_s=-2, k_{\psi}=0.7$, and $K=2$ levels of pyramid with $\sigma=8.6$ Gaussian kernel.
Due to low-speed tracking limits of the low-level linear velocity feed-forward controller on the machine, we restrict the commanded forward velocity to
$v_f\in[0.2,1]\,\si{\meter\per\second}$ during real-world experiments, and upper and lower bounds on $u_{\dot \beta}$ are $u_{\dot \beta}\in[-0.8,0.8]\,\si{\radian\per\second}$. Besides, since the linear velocity is strictly positive and the vehicle cannot turn in place, it cannot back up into an obstacle.
Therefore, we assume that only the states where $l_a \heading \cdot \nabla_{\position} \sdfSmooth (.) \leq 0 $ are relevant.
Then, abusing the formulation, $-l_s$ is considered to be equivalent to the inflation radius and is set to $\SI{2}{\meter}$ to cover the front body embodiment.

During the deployment, the wheel loader has safely operated in the scenario ($h\ge 0$) as overviewed in~\cref{fig:results:eevee}.
Hence, we conclude that despite coupling the CBF to kinematic models only, the proposed approach is appropriate for safe control onboard both a warehouse differential drive robot and an articulated wheel-loader.


\section{Conclusion}
\label{sec:conc}
We present OGM-CBF, an SDF-based CBF that is coupled to an OGM to provide safety guarantees for mobile robots in environments with arbitrarily shaped obstacles.
When the environment can only be partially observed, the OGM-CBF leverages the OGM memory to avoid a priori observed obstacles that have left the robot's field of view.
The proposed approach outperforms memory-unaware baselines that may steer into out of field of view obstacles when prompted by other hazards or the nominal control inputs. Besides, the proposed approach exploits the grid representation to consider the SDF gradient w.r.t. multiple scales of local neighborhoods, promoting steering authority in cases when the SDF gradient degenerates due to obstacle surfaces orthogonal to the heading. The OGM-CBF is sensor-agnostic, and we have utilized depth imagery, $2$D laser scanner, and $3$D LiDAR to deploy the safe control to an autonomous driving simulator, a real warehouse robot, and an articulated frame steering autonomous wheel loader, respectively. The authors suggest that future work could focus on higher-order CBFs to address complex vehicle dynamics, mitigating sensor sparsity at long range through inpainting over the grid representation, and extending the proposed framework to handle dynamic obstacles prevalent in on-road scenarios.


\bibliographystyle{ieeetr}
\bibliography{root}

@string{icra = "{IEEE} International Conference on Robotics and Automation (ICRA)"}

@string{iros = "{IEEE/RSJ} International Conference on Intelligent Robots and Systems (IROS)"}

@string{ral = "Robotics and Automation Letters"}

@string{mesas = "Modeling \& Simulation for Autonomous Systems (MESAS)"}

@string{cvpr = "{IEEE} Conference on Computer Vision and Pattern Recognition (CVPR)"}

@string{tro = "{IEEE} Transactions on Robotics"}

@string{space = "{AIAA} {SPACE}"}

@string{corl  = "Conference on Robot Learning"}

@string{cdc = "{IEEE} Conference on Decision and Control {(CDC)}"}

@string{iccma = "International Conference on Control, Mechatronics and Automation {(ICCMA)}"}

@string{iccv = "{IEEE}/{CVF} International Conference on Computer Vision {(ICCV)}"}

@string{ecc = "European Control Conference {(ECC)}"}

@string{tac = "{IEEE} Transactions on Automatic Control"}

@string{tits = "{IEEE} Transactions on Intelligent Transportation Systems"}

@string{vehits = "International Conference on Vehicle Technology and Intelligent Transport Systems {(VEHITS)}"}

@string{calcolo = "Calcolo"}

@inproceedings{toulkani2025safety,
  title={Safety Filter Design for Articulated Frame Steering Vehicles In the Presence of Actuator Dynamics Using High-Order Control Barrier Functions},
  author={Toulkani, Naeim Ebrahimi and Ghabcheloo, Reza},
  booktitle=ecc,
  pages={2108--2113},
  year={2025},
  _organization={IEEE}
}

@mastersthesis{karki2025reactive,
  title={Reactive Collision Avoidance for Articulated Frame Steering Vehicle},
  author={K{\"a}rki, Topi},
  year={2025},
  type={{M}aster's Thesis},
  school={Tampere University}
}

@inproceedings{ames2019control,
  title={Control barrier functions: {Theory} and applications},
  author={Ames, Aaron D and Coogan, Samuel and Egerstedt, Magnus and Notomista, Gennaro and Sreenath, Koushil and Tabuada, Paulo},
  booktitle=ecc,
  pages={3420--3431},
  year={2019},
  _organization={IEEE},
  doi={10.23919/ECC.2019.8796030},
}

@inproceedings{nguyen20163d,
  title={3{D} dynamic walking on stepping stones with control barrier functions},
  author={Nguyen, Quan and Hereid, Ayonga and Grizzle, Jessy W and Ames, Aaron D and Sreenath, Koushil},
  booktitle=cdc,
  pages={827--834},
  year={2016},
  doi={10.1109/CDC.2016.7798370},
}

@inproceedings{ames2014control,
  title={Control barrier function based quadratic programs with application to adaptive cruise control},
  author={Ames, Aaron D and Grizzle, Jessy W and Tabuada, Paulo},
  booktitle=cdc,
  pages={6271--6278},
  year={2014},
  doi={CDC.2014.7040372},
}

@article{wang2017safety,
  title={Safety barrier certificates for collisions-free multirobot systems},
  author={Wang, Li and Ames, Aaron D and Egerstedt, Magnus},
  journal=tro,
  volume={33},
  number={3},
  pages={661--674},
  year={2017},
  doi={10.1109/TRO.2017.2659727},
}

@article{singletary2022onboard,
title={Onboard safety guarantees for racing drones: {H}igh-speed geofencing with control barrier functions},
  author={Singletary, Andrew and Swann, Aiden and Chen, Yuxiao and Ames, Aaron D},
  journal=ral,
  volume={7},
  number={2},
  pages={2897--2904},
  year={2022},
  doi={10.1109/LRA.2022.3144777},
}

@article{ames2016control,
  title={Control barrier function based quadratic programs for safety critical systems},
  author={Ames, Aaron D and Xu, Xiangru and Grizzle, Jessy W and Tabuada, Paulo},
  journal=tac,
  volume={62},
  number={8},
  pages={3861--3876},
  year={2016},
  doi={10.1109/TAC.2016.2638961},
}

@article{xiao2023barriernet,
  title={Barrier{N}et: {D}ifferentiable control barrier functions for learning of safe robot control},
  author={Xiao, Wei and Wang, Tsun-Hsuan and Hasani, Ramin and Chahine, Makram and Amini, Alexander and Li, Xiao and Rus, Daniela},
  journal=tro,
  year={2023},
  volume={39},
  number={3},
  pages={2289-2307},
  doi={10.1109/TRO.2023.3249564},
}

@inproceedings{liu2023clf,
  title={{CLF}-{CBF} Constraints for Real-Time Avoidance of Multiple Obstacles in Bipedal Locomotion and Navigation},
  author={Liu, Jinze and Li, Minzhe and Grizzle, Jessy W and Huang, Jiunn-Kai},
  booktitle=iros,
  pages={10497--10504},
  year={2023},
  doi={10.1109/IROS55552.2023.10341626},
}

@inproceedings{han2019fiesta,
  title={{FIESTA}: {F}ast incremental {E}uclidean distance fields for online motion planning of aerial robots},
  author={Han, Luxin and Gao, Fei and Zhou, Boyu and Shen, Shaojie},
  booktitle=iros,
  pages={4423--4430},
  year={2019},
  doi={10.1109/IROS40897.2019.8968199},
}

@inproceedings{srinivasan2020synthesis,
  title={Synthesis of control barrier functions using a supervised machine learning approach},
  author={Srinivasan, Mohit and Dabholkar, Amogh and Coogan, Samuel and Vela, Patricio A},
  booktitle=iros,
  pages={7139--7145},
  year={2020},
  doi={10.1109/IROS45743.2020.9341190},
}

@article{long2021learning,
  title={Learning barrier functions with memory for robust safe navigation},
  author={Long, Kehan and Qian, Cheng and Cort{\'e}s, Jorge and Atanasov, Nikolay},
  journal=ral,
  volume={6},
  number={3},
  pages={4931--4938},
  year={2021},
  doi={10.1109/LRA.2021.3070250},
}

@inproceedings{abdi2023safe,
  title={Safe Control using Vision-based Control Barrier Function {(V-CBF)}},
  author={Abdi, Hossein and Raja, Golnaz and Ghabcheloo, Reza},
  booktitle=icra,
  pages={782--788},
  year={2023},
  doi={10.1109/ICRA48891.2023.10160805},
}

@article{blanchini1999set,
  title={Set invariance in control},
  author={Blanchini, Franco},
  journal={Automatica},
  volume={35},
  number={11},
  pages={1747--1767},
  year={1999},
  publisher={Elsevier},
  doi={10.1016/S0005-1098(99)00113-2},
}

@inproceedings{moravec1985high,
  title={High resolution maps from wide angle sonar},
  author={Moravec, Hans and Elfes, Alberto},
  booktitle=icra,
  pages={116--121},
  year={1985},
  doi={10.1109/ROBOT.1985.1087316},
}

@inproceedings{collins2007occupancy,
  title={Occupancy grid mapping: {A}n empirical evaluation},
  author={Collins, Thomas and Collins, JJ},
  booktitle={Mediterranean Conference on Control \& Automation},
  pages={1--6},
  year={2007},
  _organization={IEEE},
  doi={10.1109/MED.2007.4433772},
}

@inproceedings{wei2023surroundocc,
  title={{SurroundOcc}: {M}ulti-camera 3{D} occupancy prediction for autonomous driving},
  author={Wei, Yi and Zhao, Linqing and Zheng, Wenzhao and Zhu, Zheng and Zhou, Jie and Lu, Jiwen},
  booktitle=iccv,
  pages={21729--21740},
  year={2023},
  doi={10.1109/ICCV51070.2023.01986},
}

@article{nguyen2011stereo,
  title={Stereo-camera-based urban environment perception using occupancy grid and object tracking},
  author={Nguyen, Thien-Nghia and Michaelis, Bernd and Al-Hamadi, Ayoub and Tornow, Michael and Meinecke, Marc-Michael},
  journal=tits,
  volume={13},
  number={1},
  pages={154--165},
  year={2012},
  doi={10.1109/TITS.2011.2165705},
}

@inproceedings{li2018high,
  title={High Resolution Radar-based Occupancy Grid Mapping and Free Space Detection},
  author={Li, Mingkang and Feng, Zhaofei and Stolz, Martin and Kunert, Martin and Henze, Roman and K{\"u}{\c{c}}{\"u}kay, Ferit},
  booktitle=vehits,
  pages={70--81},
  year={2018},
  doi={10.5220/0006667300700081},
}

@inproceedings{chan2005level,
  title={Level set based shape prior segmentation},
  author={Chan, Tony and Zhu, Wei},
  booktitle=cvpr,
  volume={2},
  pages={1164--1170},
  year={2005},
  doi={10.1109/CVPR.2005.212},
}

@article{dapogny2012computation,
  title={Computation of the signed distance function to a discrete contour on adapted triangulation},
  author={Dapogny, Charles and Frey, Pascal},
  journal=calcolo,
  volume={49},
  pages={193--219},
  year={2012},
  _publisher={Springer},
  doi={10.1007/s10092-011-0051-z},
}

@article{luo2019variational,
  title={Variational analysis on the signed distance functions},
  author={Luo, Honglin and Wang, Xianfu and Lukens, Brett},
  journal={Journal of Optimization Theory and Applications},
  volume={180},
  pages={751--774},
  year={2019},
  _publisher={Springer},
  doi={10.1007/s10957-018-1414-2},
}

@inproceedings{xiao2019control,
  title={Control barrier functions for systems with high relative degree},
  author={Xiao, Wei and Belta, Calin},
  booktitle=cdc,
  pages={474--479},
  year={2019},
  doi={10.1109/CDC40024.2019.9029455},
}

@inproceedings{toulkani2022reactive,
  title={Reactive safe path following for differential drive mobile robots using control barrier functions},
  author={Toulkani, Naeim Ebrahimi and Abdi, Hossein and Koskelainen, Olli and Ghabcheloo, Reza},
  booktitle=iccma,
  pages={60--65},
  year={2022},
  _organization={IEEE},
  doi={10.1109/ICCMA56665.2022.10011466},
}

@inproceedings{dosovitskiy2017carla,
  title={{CARLA}: {A}n open urban driving simulator},
  author={Dosovitskiy, Alexey and Ros, German and Codevilla, Felipe and Lopez, Antonio and Koltun, Vladlen},
  booktitle=corl,
  pages={1--16},
  year={2017},
  _organization={PMLR},
  _url={https://proceedings.mlr.press/v78/dosovitskiy17a.html},
}

@book{rajamani2011vehicle,
  title={Vehicle Dynamics and Control},
  author={Rajamani, Rajesh},
  year={2011},
  publisher={Springer},
  address={New York, NY},
  doi={10.1007/978-1-4614-1433-9},
}

@article{li_collision-free_2025,
	title = {Collision-Free Source Seeking Control Methods for Unicycle Robots},
	volume = {70},
	_issn = {1558-2523},
	url = {https://ieeexplore.ieee.org/document/10735338/},
	doi = {10.1109/TAC.2024.3486654},
	number = {3},
	journal = tac,
	author = {Li, Tinghua and Jayawardhana, Bayu},
	year = {2025},
	pages = {2020--2027},
}

@inproceedings{bayer_cache_2020,
   title = {Speeded Up Elevation Map for Exploration of Large-Scale Subterranean Environments},
   booktitle = {{M}odelling and Simulation for Autonomous Systems ({MESAS})},
   author = {Bayer, Jan and Faigl, Jan},
   year = {2020},
   pages     = {190--202},
}

\end{document}